\documentclass[11pt]{article}

\usepackage[utf8]{inputenc} 
\usepackage[T1]{fontenc}    
\usepackage{hyperref}       
\usepackage{url}            
\usepackage{booktabs}       
\usepackage{amsfonts}       
\usepackage{nicefrac}       
\usepackage{microtype}      
\usepackage{comment}
\usepackage{xcolor}

\usepackage[utf8]{inputenc} 
\usepackage[T1]{fontenc}    
\usepackage{hyperref}       
\usepackage{url}            
\usepackage{booktabs}       
\usepackage{amsfonts}       
\usepackage{nicefrac}       
\usepackage{microtype}      

\usepackage{amsmath}

\usepackage{color}

\usepackage[pdftex]{graphicx} 
\usepackage{graphbox}
\def\O{Deepcode}
\def\o{Deepcode}

\title{\O: Feedback Codes via Deep Learning}

\author{Hyeji Kim\thanks{H. Kim, S. Oh, and P. Viswanath are with Coordinated Science Lab at the University of Illinois at Urbana Champaign. S. Oh and P. Viswanath are with the Department of Industrial and Enterprise Systems Engineering and the Department of Electrical Engineering, respectively, at the University of Illinois at Urbana Champaign. Email: \texttt{\{hyejikim,swoh,pramodv\}@illinois.edu}}, Yihan Jiang\thanks{Y. Jiang and S. Kannan are with the Department of Electrical Engineering at the University of Washington. Email: \texttt{yihanrogerjiang@gmail.com} (Y. Jiang), \texttt{ksreeram@uw.edu} (S. Kannan).}, Sreeram Kannan$^\dagger$, Sewoong Oh*, Pramod Viswanath* \ \\
University of Illinois at Urbana Champaign*, University of Washington$^\dagger$\\
}

\begin{document}

\date{}

\maketitle

\def\bh{\hat{b}}
\def\bb{\mathbf{b}}
\def\E{\mathbb{E}}
\def\yt{\tilde{y}}
\def\inn{\noindent}

\def\cb{\color{black}}

\begin{abstract}
The design of codes for communicating reliably over a  statistically well defined  channel is an important endeavor involving deep mathematical research and wide-ranging practical applications. In this work, we present the first family of codes obtained via deep learning, which significantly beats state-of-the-art codes  designed over several decades of research. The communication channel under consideration is the Gaussian noise channel with feedback, whose study was initiated by Shannon; feedback is known theoretically to improve reliability of communication, but no practical codes that do so have ever been successfully constructed. 

We break this logjam by integrating information theoretic insights harmoniously with recurrent-neural-network based encoders and decoders to create novel codes that outperform known codes by $3$ orders of magnitude in reliability. We also demonstrate several desirable properties of the codes: (a) generalization to larger block lengths, (b) composability with known codes, (c) adaptation to practical constraints. This result also has broader ramifications for coding theory: even when the channel has a clear mathematical model, deep learning methodologies, when combined with channel-specific information-theoretic insights, can potentially beat state-of-the-art codes constructed over decades of mathematical  research.

\end{abstract}

\section{Introduction}\label{intro}

The ubiquitous digital communication enabled via wireless (e.g. WiFi, mobile, satellite) and wired  (e.g. ethernet, storage media, computer buses) media has been the plumbing underlying the current information age.
The advances of reliable and efficient digital communication have been primarily driven by the design of codes which allow the receiver to recover messages reliably and efficiently under noisy conditions.
The discipline of coding theory has made significant progress in the past seven decades since Shannon's celebrated work in 1948~\cite{shannon1948mathematical}.
As a result, we now have near optimal codes in a canonical setting, namely, additive white Gaussian noise (AWGN) channel.
However, several channel models of great practical interest lack efficient and practical coding schemes.



A channel with {\em feedback} (from the receiver to the transmitter) is an example of a long-standing open problem and with significant  practical importance. Modern wireless communication includes feedback in one form or the other; for example, the feedback can be the received value itself, or quantization of the received value or an automatic repeat request (ARQ)~\cite{ChoiJSKL10}.
%
 %
Accordingly, there are different models for channels with feedback, and among them, 
 the AWGN channel with \emph{output} feedback is a model that captures the essence of channels with feedback; this model is also classical,
 introduced by Shannon in 1956~\cite{shannon1956zero}. In this channel model, the received value is fed back (with unit time delay) to the transmitter without any processing (refer to Figure~\ref{fig0} for an illustration of channel).
  Designing codes for this channel via deep learning approaches is the central focus of this paper.


While the output feedback does not improve the Shannon capacity of the AWGN channel \cite{shannon1956zero}, it is known to provide better reliability at finite block lengths \cite{schalkwijk1966coding}.  On the other hand,  practical coding schemes have not been successful in harnessing the feedback gain thereby significantly limiting the use of feedback in practice. This state of the  art is at odds with the theoretical predictions of the gains in reliability via using feedback:
the seminal work of Schalkwijk and Kailath~\cite{schalkwijk1966coding} proposed  a (theoretically) achievable scheme  (S-K scheme) with superior reliability guarantees, but which suffers from extreme sensitivity to both the precision of the numerical computation and noise in the feedback~\cite{Schalkwijk_part2,Gallager--Nakiboglu2010}.
Another competing scheme of~\cite{Chance--Love2011} is designed for channels with noisy feedback, but  not only is the reliability poor, it  is almost independent of the feedback quality, suggesting that the feedback data is not being fully exploited.
More generally, it has been proven that no {\em linear} code incorporating the noisy output feedback can perform well \cite{kim2007gaussian}. 
This is especially troubling since all practical codes are linear and linear codes are known to achieve capacity (without feedback) \cite{elias1955coding}.

In this paper, 
 we demonstrate new neural network-driven encoders (with matching decoders) that operate significantly better (100--1000 times) than state of the art on the AWGN channel with (noisy) output feedback.
We show that architectural insights from simple communication channels with feedback, when coupled with recurrent neural network architectures, can discover novel codes.   We consider Recurrent Neural Network (RNN) parameterized encoders (and decoders), which are inherently {\em nonlinear} and map information bits {\em directly} to real-valued transmissions in a sequential manner.

This is the first family of codes obtained via deep learning which beats state-of-the-art codes, signaling a potential shift in code design, which historically has been driven by individual human ingenuity with sporadic progress over the decades. {{\cb}Henceforth, we call this new family of codes \o.}\ We also note that although there has been significant recent interest in deep-learning driven designs of decoders for known codes, or end-to-end designs with varying degrees of success~\cite{felix2018ofdm,cammerer2018end,o2017introduction,ibnkahla2000applications,o2016learning,kim2018communication,nachmani2016learning,nachmani2018deep,tan2018improving,over_the_air,endtoend_nomodel,jointsc,Cho_OFDM},  none of these works are able to design novel codes that can beat the state of the art. 
{{\cb}In a different context, for distributed computation, where the encoder adds redundant computations so that the decoder can reliably approximate the desired computations under unavailabilities,~\cite{Kosaian18} showed that neural network based codes can beat the state-of-the-art codes.} 
 We also demonstrate the superior performance of variants of \o\ under a variety of practical constraints. Furthermore, \o\ 
has complexity comparable to traditional codes, even without any effort at optimizing the storage and run-time complexity of the neural network architectures. Our main contributions are as follows:
 \begin{enumerate}
\item We demonstrate \o\ -- a new family of RNN-driven neural codes that have {\em three orders of magnitude} better reliability than state of the art with both noiseless and noisy feedback. Our results are significantly driven by the intuition obtained from information and coding theory, in designing a series of progressive improvements in the neural network architectures (Section~\ref{sec3} and~\ref{sec4}).

\item We show that variants of \o\ 
significantly outperform state-of-the art codes even under a variety of practical constraints (example: delayed feedback, very noisy feedback link) (Section~\ref{sec4}). 
\item 
We show {\em composability}: 
\o\ naturally concatenates with a traditional inner code and demonstrates continued improvements in reliability as the block length increases (Section~\ref{sec4}).
\item Our interpretation and analysis of \o\ 
 provide guidance on {{\cb}the fundamantal} 
 understanding of how the feedback {{\cb}can} be used and some information theoretic insights into designing codes for channels with feedback (Section~\ref{sec5}).
\item {{\cb}We discuss design decisions and demonstrate practical gains of \o\ 
in practical cellular communication systems (Section~\ref{sec6}).} 
\end{enumerate}

\section{Problem formulation}\label{sec2}


The most canonical channel studied in the literature (example: textbook material \cite{cover2012elements})
and also used in modeling practical scenarios (example: 5G LTE standards) is
the additive white Gaussian noise (AWGN) channel {\em without} feedback.
Concretely, the encoder takes in $K$ information bits jointly,
$\bb = (b_1, \cdots, b_K)\in\{0,1\}^K$,
and outputs $n$ real valued signals
to be transmitted over a noisy channel (sequentially).
At the $i$-th transmission for each $i\in \{1,\ldots,n\}$, a transmitted symbol $x_i\in{\mathbb R}$ is
corrupted by an independent Gaussian noise $n_i \sim \mathcal{N}(0,\sigma^2)$,
and the decoder receives $y_i = x_i + n_i \in {\mathbb R}$.
After receiving the $n$ received symbols,
the decoder makes a decision on which  information bit $\bb$ was sent,
out of $2^K$ possible choices. The goal is to maximize the probability of correctly decoding the received symbols
and recover $\bb$.

Both the encoder and the decoder are functions, mapping
$\bb\in\{0,1\}^K$ to $\mathbf{x} \in {\mathbb R}^n$
and $\mathbf{y}\in {\mathbb R}^n$ to $\mathbf{\bh} \in \{0,1\}^K$, respectively.
The design of a good code (an encoder and a corresponding decoder) addresses both
$(i)$ the statistical challenge of achieving a small error rate;
and $(ii)$ the computational challenge of achieving the desired error rate with efficient encoder and decoder.
Almost a century of progress in this domain of
 coding theory has produced
 several innovative codes
 that efficiently achieve small error rate,
 including convolutional codes, Turbo codes, LDPC codes,
 and polar codes. These codes are known to perform close to the  fundamental limits on reliable communication \cite{polyanskiy2010channel}.


\begin{figure}[!ht]
\vspace{-1em}
	\centering
	\includegraphics[width=0.5\textwidth]{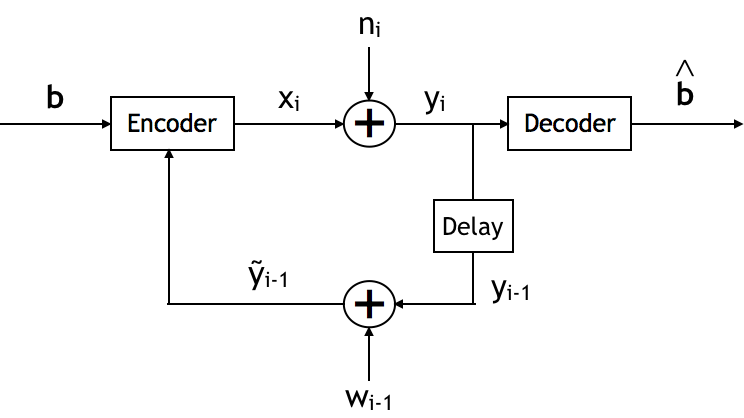}%
	\caption{AWGN channel with noisy output feedback}
	\label{fig0}
	\vspace{-0.5em}
\end{figure}

In a canonical AWGN channel {\em with} noisy feedback,
the received symbol $y_i$ is transmitted back to the encoder after one unit time of delay
and via another additive white Gaussian noise {\em feedback channel} (Figure~\ref{fig0}).
The encoder can use this feedback symbol to sequentially and adaptively decide
what symbol to transmit next.
At time $i$ the encoder receives a noisy view of what was received at the receiver (in the past by one unit time),
$\yt_{i-1} = y_{i-1} + w_{i-1}\in{\mathbb R}$, where the noise is independent and distributed as
$w_{i-1} \sim \mathcal{N}(0,\sigma_F^2) $.
Formally, an {\em encoder} is now a function that sequentially maps the information bit vector  $\bb$
and the feedback symbols $\yt^{i-1}_1 = (\yt_1,\cdots, \yt_{i-1})$ received thus far
to a transmit symbol $x_i$:
$f_i \ \colon\ \  (\bb,\yt_1^{i-1}) \mapsto x_i, \ \;\;\; i \in\{ 1,\cdots, n\}\;$
and a \emph{decoder} is a function that
 maps the received sequence $y_1^n=(y_1,\cdots,y_n)$
 into estimated information bits:
$g \ \colon \ \ y_1^n \mapsto \mathbf{\bh} \in \{0,1\}^K$.

%
%
%
The standard measures of performance are
the average bit error rate (BER)
 defined as
 ${\rm BER} \equiv (1/K) \sum_{i=1}^K {\mathbb P} (b_i \neq \bh_i)$ {\color{black} and the block error rate (BLER) defined as ${\rm BLER} \equiv {\mathbb P} (\mathbf{b} \neq \mathbf{\bh})$,}
 where the randomness comes from the forward and feedback channels and
 any other sources of randomness
 that might be used in the encoding and decoding processes. 
It is  standard (both theoretically and practically) to have an
average power constraint, i.e., $(1/n) \E[\mathbf{\| x\| }^2] \le 1$,
where ${\mathbf x}=(x_1,\cdots,x_n)$
and the expectation is over the randomness in choosing the information bits $\bb$ uniformly at random,
the randomness in the noisy feedback symbols $\yt_i$'s,
and any other randomness used in the encoder.

\begin{figure}[!ht]
\vspace{-1em}
	\centering
	\includegraphics[width=0.4\textwidth]{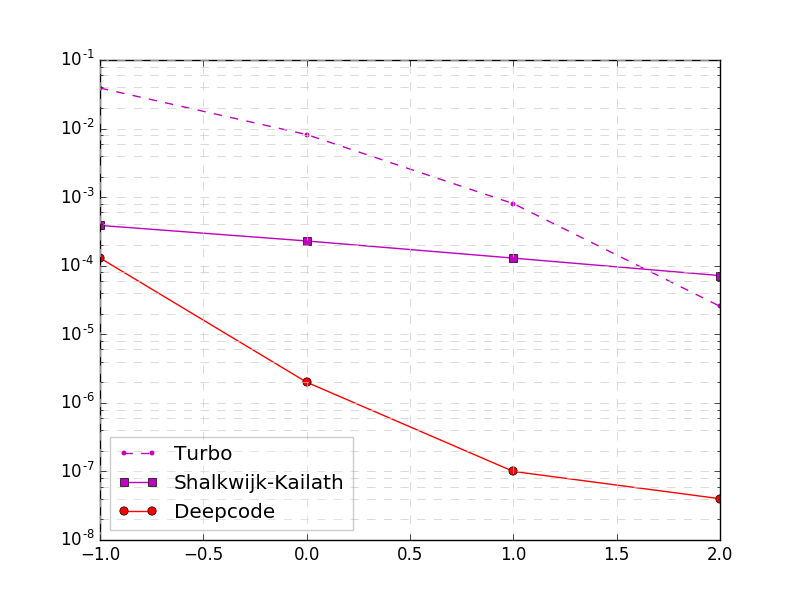}\ \ \ \ \ \ \ \ \ \ \ \ \ \ \ \ \ \ \ 
	\put(-207,80){BER}
	\put(-135,-3){SNR = $-10 \log_{10} \sigma^2$}\ \ \ \ \ \ 
		\includegraphics[width=0.4\textwidth]{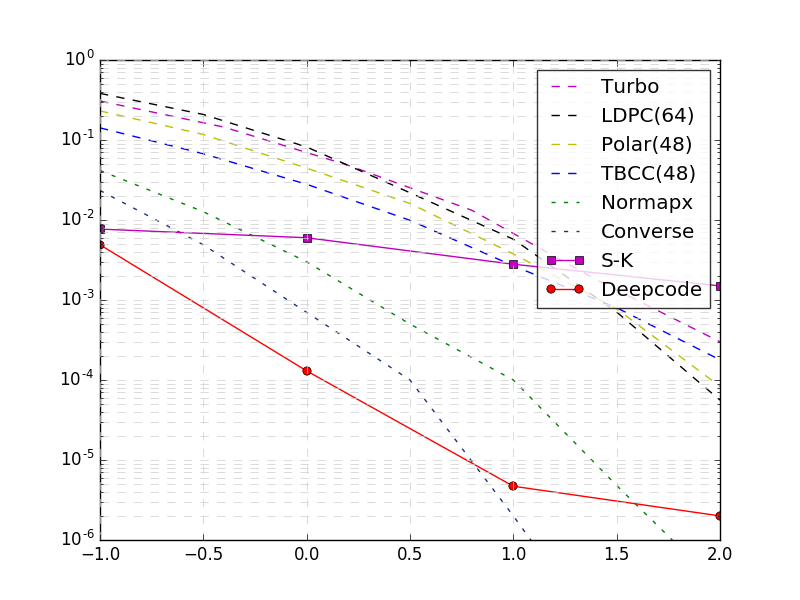}
	\put(-207,80){BLER}
	\put(-135,-3){SNR = $-10 \log_{10} \sigma^2$}
	\caption{
	\O\ 
	significantly outperforms the baseline of S-K  and Turbo code,
		on block-length 50 and noiseless feedback in BER (left) {\color{black}and BLER (right). \O\ 
		also outperforms all  state-of-the art codes (without feedback) in BLER (right).}}
	\label{fig1}
	\vspace{-0.5em}
\end{figure}

While the capacity of the channel remains the same in the presence of feedback \cite{shannon1956zero}, the reliability can increase significantly as demonstrated by the  celebrated result of
Schalkwijk and Kailath (S-K),
\cite{schalkwijk1966coding}, which is described in detail in Appendix~\ref{sec:CLSK}.
Although the  optimal {theoretical} performance is met by the S-K scheme,
critical drawbacks make it fragile. 
Theoretically, the scheme critically relies on exactly noiseless feedback (i.e.~$\sigma_F^2=0$),
and does not  extend to channels with even arbitrarily small amount of noise in the feedback  (i.e.~$\sigma_F^2>0$).
Practically, the scheme is extremely sensitive to  numerical precisions; we see this in  Figure \ref{fig1}, 
where the numerical errors dominate the performance of the S-K scheme,
with a practical choice of
 MATLAB implementation with a precision of 16 bits to represent floating-point numbers.

Even with a noiseless feedback channel with $\sigma_F^2=0$,
which the S-K scheme is designed for,
it is outperformed  significantly by our proposed \o\ 
(described in detail in Section \ref{sec3}).
At moderate SNR of 2 dB, \o\ 
 can outperform S-K scheme by three orders of magnitude in BER.
In Figure~\ref{fig1} (left), the resulting BER is shown
as a function of the Signal-to-Noise Ratio (SNR)
defined as $-10 \log_{10} \sigma^2$, where we consider the setting of rate $1/3$ and information block length of $K=50$ (hence, $n=150$). 
Also shown as a baseline is
an LTE turbo code
which does not use any feedback.
\O\ 
 exploits the feedback symbols to achieve a significant gain of two orders of magnitude consistently over the Turbo code for all SNR.
{\color{black}In Figure~\ref{fig1} (right), BLER of \o\ 
is shown
as a function of the Signal-to-Noise Ratio (SNR), 
together with state-of-the art polar, LDPC, and tail-bitting convolutional codes (TBCC) in a 3GPP document for the 5G meeting~\cite{5G_doc} (we refer to Appendix~\ref{5G_doc} for the details of these codes used in the simulation). 
\O\ 
significantly improves over all state-of-the-art codes of similar block-length and the same rate.  
Also plotted as a baseline 
are the theoretically estimated performance of the best code with no efficient decoding schemes. 
This impractical baseline lies between approximate achievable BLER (labelled Normapx in the figure) 
and a converse to the BLER (labelled Converse in the figure) from ~\cite{polyanskiy2010channel, Tomaso}. 
}
We note that there are  schemes proposed more recently that 
address the sensitivity to noise in the feedback, a major drawback of the S-K scheme. 
However, these schemes either still suffer from similar sensitivity to numerical precisions at the decoder
\cite{Kim--Lapidoth--Weissman2007},
or are incapable of exploiting the feedback information
\cite{Chance--Love2011}
as we illustrate in Figure \ref{noisy4} in experiments with noisy feedback.

\section{\O: neural encoder and decoder}\label{sec3}

A natural strategy to create a feedback code is to utilize a recurrent neural network (RNN) as an encoder since $(i)$ communication with feedback is naturally a sequential process and $(ii)$ we can exploit the sequential structure for efficient decoding. We propose representing the encoder and the decoder as RNNs,  training them jointly under AWGN channels with noisy feedback, and  minimizing the error in decoding the information bits. However, in our experiments, we find that this strategy by itself is insufficient to achieve any performance improvement with feedback.

We exploit  information theoretic insights to enable improved performance, by considering  the erasure channel with feedback:  here transmitted bits are either received perfectly or erased, and whether the previous bit was erased or received perfectly is fed back to the transmitter. In such a channel, the following two-phase scheme can be used: transmit a block of symbols, and then transmit whichever symbols were erased in the first  block (and ad infinitum). This motivates a two-phase scheme, where uncoded bits are sent in the  first phase, and then based on the feedback in the first phase, coded bits are sent in the second phase; thus the code only needs to be designed for the second phase. Even inside this two-phase paradigm, several architectural choices need to be made.
We show in this section that
 these  intuitions can be critically employed  to innovate neural network  architectures.

Our experiments focus on the setting of rate 1/3 and information block length of 50 for concreteness.\footnote{Source codes are available under \url{https://github.com/hyejikim1/Deepcode} (Keras) and \url{https://github.com/yihanjiang/feedback_code} (PyTorch)} That is, the encoder
maps $K=50$ message bits to a codeword of length
$n=150$.
We discuss generalizations to longer block lengths in Section~\ref{sec4}.
\begin{figure}[!h]
\centering
\vspace{-1em}
	\includegraphics[width=0.4\textwidth]{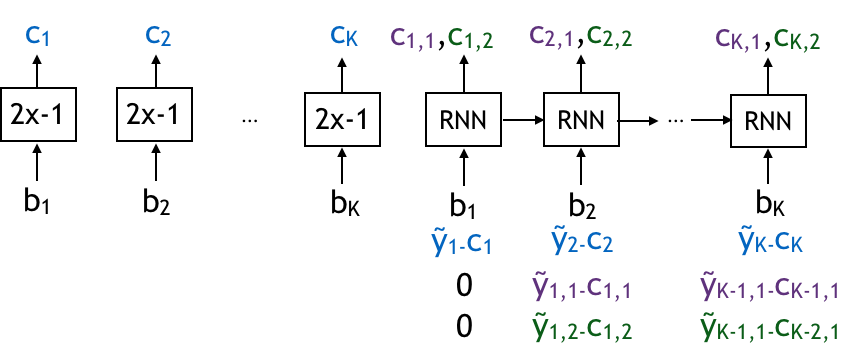}
	\put(-170,80){Encoder A: RNN feedback encoder}
	\hspace{0.6cm}
	\includegraphics[width=0.45\textwidth]{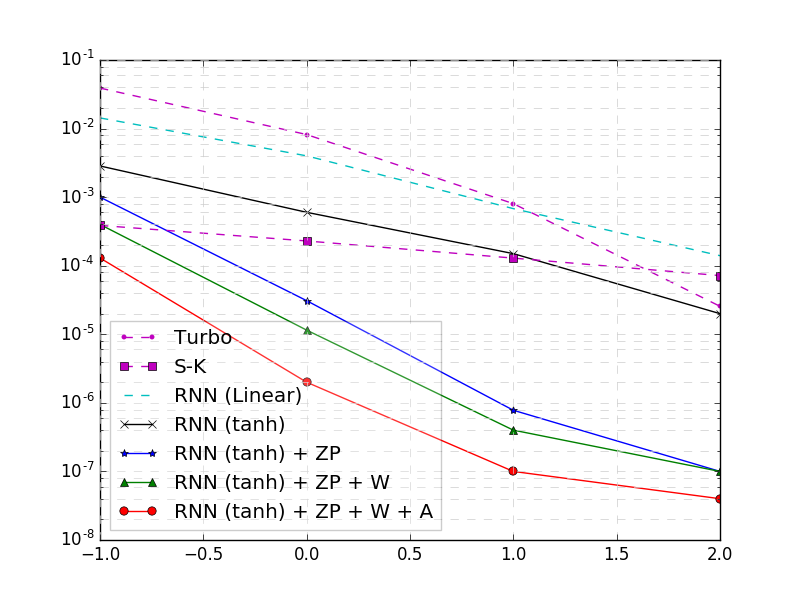}
	\put(-225,90){BER}
	\put(-141,-3){SNR =$-10 \log_{10} \sigma^2$}
	\caption{
		Building upon a simple linear RNN encoder (left), we progressively improve the architecture.
		Eventually with RNN(tanh)+ZP+W+A architecture formally described in Section
		\ref{sec3},  we significantly outperform the baseline of S-K scheme and Turbo code,
		by several orders of magnitude in the bit error rate,
		on block-length 50 and noiseless feedback ($\sigma_F^2=0$).}\label{figure1}
\end{figure}
%
%

\bigskip
\noindent {\bf A. RNN feedback encoder/decoder (RNN (linear) and RNN (tanh)).}  We propose an encoding scheme that progresses in two phases.
 In the first phase, the $K$ information bits are sent raw (uncoded) over the AWGN channel. In the second phase, $2K$ coded bits are generated based on the information bits $\bb$
 and (delayed) output feedback and sequentially transmitted.
 We propose a decoding scheme using two layers of bidirectional gated recurrent units (GRU).
When jointly trained, a {\em linear} RNN encoder achieves performance close to Turbo code that does not use the feedback information at all as shown in  Figure \ref{figure1}. 
(To generate plots in Figure~\ref{figure1}, we take an average bit error rate over $10^8$ bits for SNR = $-1,0$dB and $10^9$ bits for SNR= $1,2$dB.)
With a {\em non-linear} activation function of $\tanh(\cdot)$,
the performance improves, achieving BER close to the existing S-K scheme.
Such a gain of non-linear codes over linear ones is in-line with
 theory \cite{Kim--Lapidoth--Weissman2007}.


\bigskip
\inn \textbf{Encoding.}
 The architecture of the encoder is shown in Figure~\ref{figure1}.
The encoding process has two phases. In the first phase, the encoder simply transmits the $K$ raw message bits. That is, the encoder maps $b_k$ to $c_k = 2b_k-1$ for $k \in \{1,\cdots,K\}$, and
stores the feedback $\tilde{y}_1, \cdots, \tilde{y}_K$ for later use.
In the second phase, the encoder generates a coded sequence of length $2K$
(length $(1/r-1)K$ for general rate $r$ code) through a single directional RNN. In particular, each $k$-th RNN cell generates two coded bits $c_{k,1},c_{k,2}$ for $k\in\{1,\ldots,K\}$, which uses both the information bits and (delayed) output feedback from the earlier raw information bit transmissions. The input to the $k$-th RNN cell is of size four: $b_k$, $\tilde{y}_k - c_k$ (the estimated noise added to the $k$-th message bit in phase 1) and the most recent two noisy feedbacks from phase 2: $\tilde{y}_{k-1,1} - c_{k-1,1}$ and $\tilde{y}_{k-1,2} - c_{k-1,2}$. 
Note  that we use $\tilde{y}_{k,j} = c_{k,j} + n_{k,j}+w_{k,j}$ to denote the feedback received from the transmission of $c_{k,j}$ for
$k\in\{1,\cdots,K\}$ and $j\in\{1,2\}$, and
$n_{k,j}$ and $w_{k,j}$ are corresponding forward and feedback channel noises, respectively.

To generate codewords that satisfy power constraint, we put a normalization layer to the RNN outputs so that each coded bit has a mean $0$ and a variance $1$. During training, the normalization layer subtracts the batch mean from the output of RNN and divides by the standard deviation of the batch. After training, we compute the mean and the variance of the RNN outputs over $10^6$ examples. In testing, we use the precomputed means and variances.
Further details on the encoder are shown in Appendix \ref{schemeA}.

\bigskip
\inn \textbf{Decoding.}
Based on the received sequence $\mathbf{y} = (y_1,\cdots,y_k, y_{1,1},y_{1,2},y_{2,1},y_{2,2}, \cdots, y_{K,1},y_{K,2})$ of length $3K$, the decoder estimates $K$ information bits.
For the decoder, we use a two-layered bidirectional GRU, where the input to the $k$-th GRU cell is a tuple of three received symbols, $(y_k, y_{k,1}, y_{k,2})$.
We refer to Appendix \ref{schemeA} for detailed description of the decoder.

\bigskip
\inn \textbf{Training.}
Both the encoder and decoder are trained {\em jointly} using binary cross-entropy as the loss function over $4 \times 10^6$ examples, with batch size 200. 
The input to the neural network is $K$ information bits and the output is $K$ estimated bits (as in the autoencoder setting). AWGN channels are simulated for the channels from the encoder to the decoder and from decoder to the encoder.
In training, we set the forward SNR to be test SNR and feedback SNR to be the test feedback SNR. We randomly initialize weights of the encoder and the decoder. We observed that training with random initialization of encoder-decoder gives a better encoder-decoder compared to initializing with a pre-trained encoder/decoder by sequential channel codes for non-feedback AWGN channels (e.g. convolutional codes). We also use a decaying learning rate and gradient clipping; we reduce the learning rate by 10 times after training with $10^6$ examples, starting from $0.02$. Gradients are clipped to 1 if $L_2$ norm of the gradient exceeds 1, preventing gradients from becoming too large. 

\begin{figure}[!ht]
\centering
\includegraphics[width=0.31\textwidth]{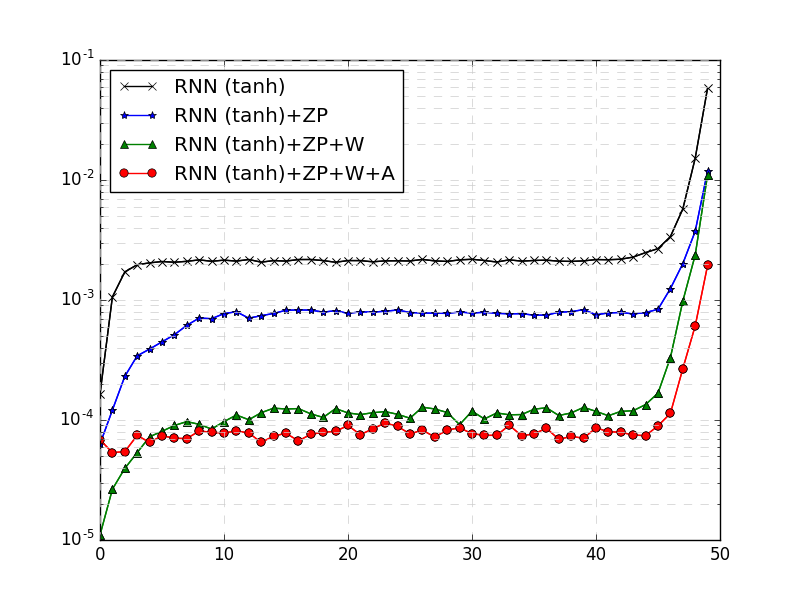}
	\put(-155,105){BER of $b_k$}
	\put(-106,-5){Position ($k$)}
\includegraphics[width=0.31\textwidth]{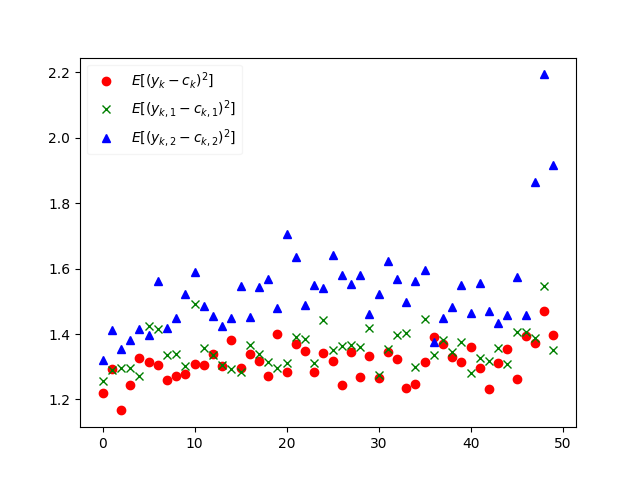} 
	\put(-152,102){\ \ \ \ $\sigma^2$}
	\put(-106,-3){Position ($k$)}
\includegraphics[width=0.31\textwidth]{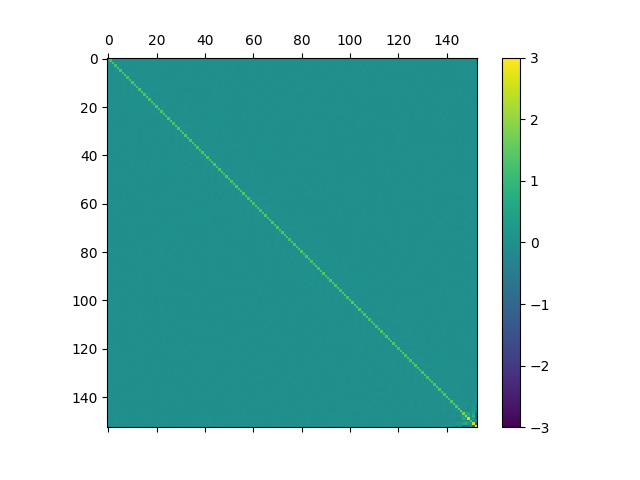} 
\caption{(Left) A naive RNN(tanh) code gives a high BER in the last few information bits. With the idea of zero padding and power allocation,  the RNN(tanh)+ZP+W+A architecture gives a BER that varies less across the bit position, and overall BER is significantly improved over the naive RNN(tanh) code. (Middle) Noise variance across bit position which results in a block error: High noise variance on the second parity bit stream $(c_{1,2},\cdots,c_{K,2})$ causes a block error. (Right) Noise covariance: Noise sequence which results in a block error does not have a significant correlation across position.
}\label{ber_pos} 
\end{figure}

\bigskip
\inn \textbf{Typical error analysis.}  
Due to the recurrent structure in generating coded bits $(c_{k,1},c_{k,2})$, 
the coded bit stream carries more information on the first few bits than the last few bits (e.g. $b_1$ than $b_K$). This results in more errors in the last information bits, as shown in Figure~\ref{ber_pos}, where we plot the average BER of $b_k$ for $k=\{1,\cdots,K\}$.

\bigskip
\noindent {\bf B. RNN feedback code with zero padding (RNN (tanh) + ZP).}
In order to reduce high errors in the last information bits, as shown in Figure~\ref{ber_pos}, we apply the zero padding (ZP) technique; we pad a zero in the end of information bits, and transmit a codeword for the padded information bits (see Appendix~\ref{schemeB} for implementation details). 
By applying zero padding, the BER of the last information bits, as well as other bits, drops significantly, as shown in Figure~\ref{ber_pos}.
Zero padding requires a few extra channel usages (e.g. with one zero padding, we map 50 information bits to a codeword of length 153). However, due to the significant improvement in BER, it is widely used in sequential codes (e.g. convolutional codes and turbo codes).
\bigskip
\inn \textbf{Typical error analysis.} 
To see if there is a pattern in the noise sequence which makes the decoder fail, 
we simulate the code and the channels and look at the first and second order noise statistics which result in the decoding error. 
In Figure~\ref{ber_pos} (Middle), we plot the average variance of noise added to $b_k$ in the first phase and $c_{k,1}$ and $c_{k,2}$ in the second phase, as a function of $k$, which results in the (block) error in decoding. From the figure, we make two observations: ($i$) large noise in the last bits causes an error, and ($ii$) large noise in $c_{k,2}$ is likely to cause an error, which implies that
 the raw bit stream and the coded bit streams are not equally robust to the noise -- an observation that will be exploited next. In Figure~\ref{ber_pos} (Right), we plot noise covariances that result in a decoding error. From Figure~\ref{ber_pos} (Right), we see that there is no particular correlation within the noise sequence that makes the decoder fail. 
\bigskip
\noindent {\bf C. RNN feedback code with power allocation (RNN(tanh) + ZP + W).}
Based on the observation that the raw bit $c_{k}$ and coded bit $c_{k,1}, c_{k,2}$ are not equally robust, as shown in Figure~\ref{ber_pos} (Middle),
we introduce trainable weights which allow allocating different amount of power to the raw bit stream and coded bit streams (see Appendix~\ref{schemeC} for implementation details). By introducing and training these weights, we achieve the improvement in BER as shown in Figures~\ref{figure1} and~\ref{ber_pos}.

\bigskip
\inn \textbf{Typical error analysis.} While the average BER is improved by about an order of magnitude for most bit positions as shown in Figure~\ref{ber_pos} (Left), the BER of the last bit remains about the same. On the other hand, the BER of the first few bits is now smaller, suggesting the following bit-specific power allocation method.


\bigskip
\noindent {\bf D.\ \O: RNN feedback code with bit power allocation (RNN(tanh)\!+\!ZP\!+\!W\!+\!A).} 
One way to resolve the unbalanced error according to bit position is to use power allocation. Ideally, we would like to reduce the power for the first information bits and increase the power for the last information bits so that we help transmission of the last few information bits more than the first information bits. However, it is not clear how much power to allow for the first few information bits and the last few information bits. Hence, we introduce a weight vector allowing the power of bits in different positions to be different (as illustrated in Figure~\ref{enc4} in Appendix~\ref{schemeD}). The resulting BER curve is shown in Figure~\ref{figure1}. We can see that the BER is noticeably decreased. In Figure~\ref{ber_pos}, we can also see that the BER in the last bits is reduced, and we can also see that the BER in the first bits is increased, as expected.
Our use of unequal
power allocation across information bits is in-line with other approaches from information/coding theory~\cite{Duman1997},~\cite{irregular_pw}. {{\cb}We call this neural code \o}.

\bigskip
\inn \textbf{Typical error analysis.}
As shown in Figure~\ref{ber_pos}, the BER at each position remains about the same except for the last few bits.
This suggests a symmetry in our code and nearest-neighbor-like decoder.
For an AWGN channel without feedback, it is known that the optimal decoder (nearest neighbor decoder) under a symmetric code (in particular, each coded bit follows a Gaussian distribution) is robust to the distribution of noise~\cite{Lapidoth1996}; the BER does not increase if we keep the power of noise and only change the distribution. As an experiment demonstrating the robustness of \o, 
 in Appendix~\ref{app:robustness}, we show that BER of \o\ 
 does not increase if we keep the power of noise and change the distribution from i.i.d. Gaussian to bursty Gaussian noise. 



\bigskip
\noindent {\bf Complexity.} Complexity and latency, as well as reliability, are important metrics in practice, as the encoder and decoder need to run in real time on mobile devices.
\O\ 
has computational complexity and latency comparable to currently used codes (without feedback) that are already in communication standards.
Turbo decoder, for example, is a belief-propagation decoder with many (e.g., 10 -- 20) iterations, and each iteration is followed by a permutation. Turbo encoder also includes a permutation of information bits (of length $K$). On the other hand, the proposed neural encoder in \o\ is a single layered RNN encoder with 50 hidden units, and the neural decoder in \o\  is a 2-layered GRU decoder, also with 50 hidden units, all of which are matrix multiplications that can be parallelized. Ideas such as  knowledge distillation \cite{hinton2015distilling} and  network binarization \cite{rastegari2016xnor} can be used to potentially further reduce the complexity of the network.

\section{Practical considerations: noise, delay, coding in feedback, and blocklength}\label{sec4}
We considered so far the AWGN channel with noiseless output feedback with a unit time-step delay. 
In this section,
we demonstrate the robustness of \o\ 
 (and its variants) 
under two variations on the feedback channel, \emph{noise} and \emph{delay}, 
and present generalization to longer block lengths. We show 
%
that
 ($a$)
\o\ 
 and its variant that allows a $K$-step delayed feedback
are more reliable than the state-of-the-art schemes in channels with \emph{noisy} feedback; 
 ($b$) by allowing the receiver to feed back an RNN encoded output instead of its raw \emph{output}, and learning this RNN encoder, 
 we achieve a further improvement in reliability, demonstrating the power of encoding in the feedback link; ($c$)
 \o\ 
 concatenated with turbo code achieves superior error rate decay as block length increases with noisy feedback. 

\begin{figure}[!ht]
\centering
\includegraphics[width=0.33\textwidth]{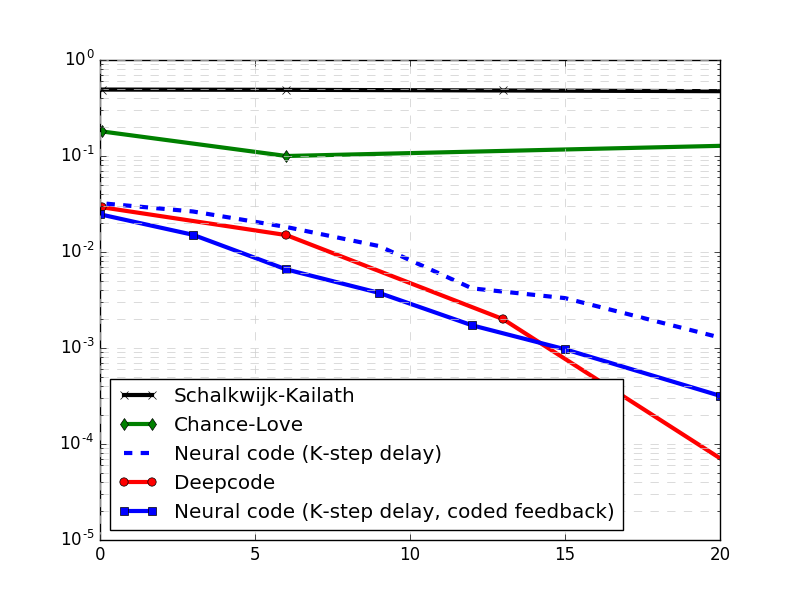}
	\put(-157,109){BER}
	\put(-132,-6){SNR of feedback channel}
\includegraphics[width=0.33\textwidth]{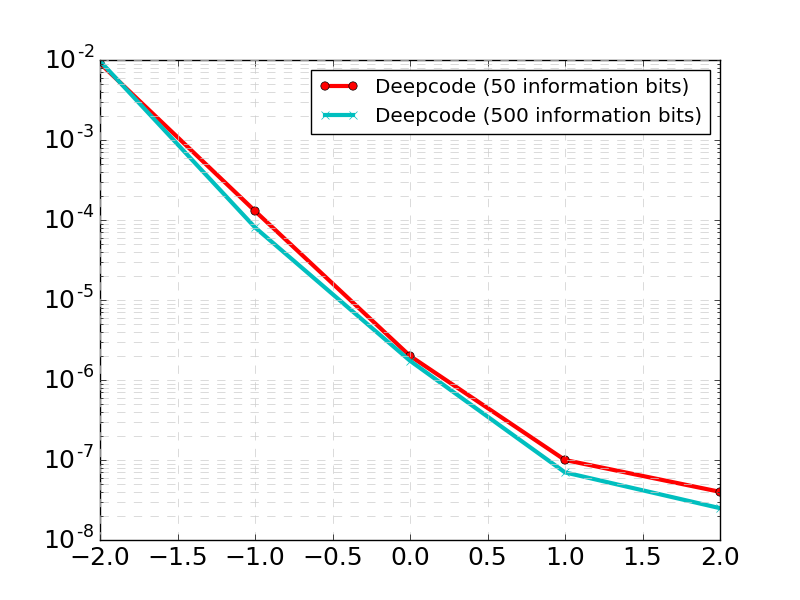}
	\put(-157,109){BER}
	\put(-90,-6){SNR}
\includegraphics[width=0.33\textwidth]{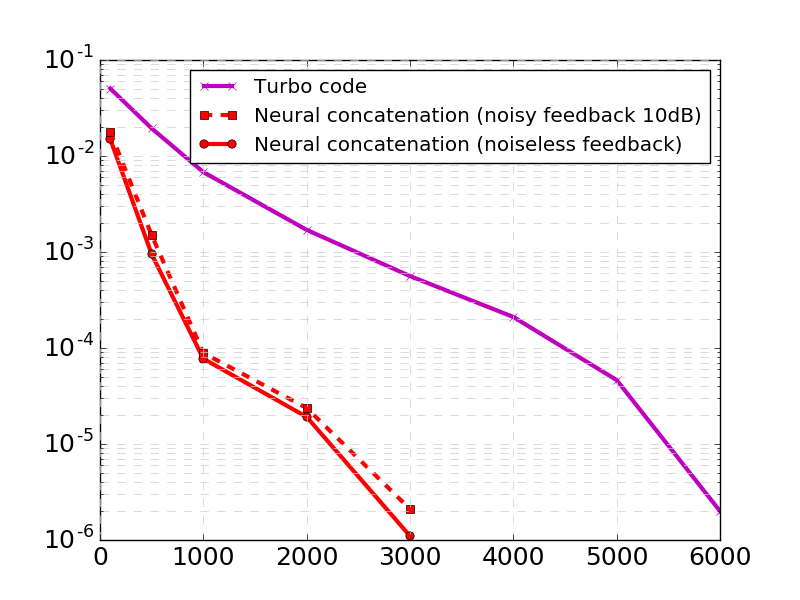}
	\put(-157,109){BER}
	\put(-105,-6){Blocklength}
\caption{(Left)
\O\  (introduced in Section~\ref{sec3}) and its variant code that allows $K$ time-step delay significantly outperform the two baseline schemes in noisy feedback scenarios. 
Another variant of \o\ 
which allows the receiver to feed back an \emph{RNN encoded} output (with $K$-step delay) performs even better than \o\ 
with \emph{raw} output feedback (with unit-delay), demonstrating the power of coding in the feedback.
 (Middle)
 By unrolling the RNN cells of \o, the BER of \o\ 
remains unchanged for block lengths 50 to 500. 
 (Right) Concatenation of \o\ and turbo code (with and without noise in the feedback) achieves BER that decays exponentially as block length increases, faster than turbo codes (without feedback) at the same rate. 
 }\label{noisy4}
\end{figure}
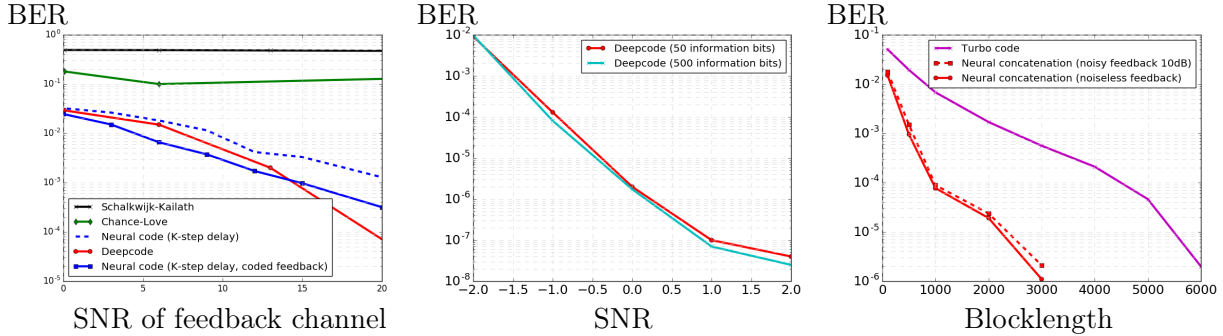

\bigskip
\noindent {\bf Noisy feedback.} We show that \o
, trained on AWGN channels with \emph{noisy} output feedback, achieves a significantly smaller BER than both S-K and C-L schemes \cite{Chance--Love2011} in AWGN channels with \emph{noisy} output feedback.
In Figure~\ref{noisy4} (Left), we plot the BER as a function of the feedback SNR for S-K scheme, C-L scheme, and \o\ 
for a rate 1/3 code with 50 information bits, where we fix the forward channel SNR to be 0dB. As feedback SNR increases, 
 we expect the BER to decrease. However, as shown in Figure~\ref{noisy4} (Left), both C-L scheme, designed for channels with noisy feedback, and S-K scheme  
 are sensitive to even a small amount of noise in the feedback, and reliability is almost independent of feedback quality. 
 
 \O\ 
outperforms these two baseline (linear) codes by a large margin, with decaying error as feedback SNR increases, showing that \o\ 
harnesses \emph{noisy} feedback information to make communication more reliable. This is highly promising as the performance with noisy feedback is directly related to the practical communication channels.
  To achieve the performance shown in  Figure~\ref{noisy4}, for example the line in red, 
 training with matched SNR is required. 
 For each datapoint, we use different neural codes specifically trained at the same SNR at the test noise. 
 In Section \ref{sec5}, we discuss how \o\ 
 differs depending on what SNR it was trained on, hence it is not universal. 
 

\bigskip
\noindent {\bf Noise feedback with delay.}
We model the practical constraint of \emph{delay} in the feedback,  
by introducing a variant of \o\ 
that works with a $K$ time-step delayed feedback (discussed  in detail in Appendix~\ref{sec:delayedfb}); recall $K$ is the number of information bits and this code tolerates a large delay in the feedback. 
We see from Figure~\ref{noisy4} (Left), 
that these neural codes 
are robust against delay in the feedback for noisy feedback channels of SNR up to 12dB.

\bigskip
\noindent {\bf Noisy feedback with delay and coding.} 
It is natural to allow the receiver to send back a general {\em function} of its past received values, 
i.e., receiver encodes its output and sends the coded and real-valued symbol.  
Designing the code for this setting is challenging as it involves designing 
two encoders (one at the transmitter and another at the receiver) and 
one decoder (at the receiver) jointly in a sequential manner. 
Further, both the transmitter and the receiver are performing encoding and decoding simultaneously and sequentially. 
We propose using RNN as a receiver encoder that maps noisy channel output 
to the transmitted feedback, with implementation details in Appendix~\ref{sec:delayedfb}.
Figure~\ref{noisy4} demonstrates the 
feedback coding gain over \o\ 
with uncoded feedback. 
When feedback channel is less noisy, neural codes with coded feedback and  
$K$-step delays can even outperform the neural code with uncoded feedback with one-step delays, 
overcoming the challenges  of  larger delays in practice. 



\bigskip
\noindent {\bf Generalization to longer block lengths.} In wireless communications, a wide range of blocklengths are of interest (e.g., 40 to 6144 information bits in LTE standards). In previous sections, we considered block length of $50$ information bits. Here we show how to  generalize \o\ 
to longer block lengths and achieve an improved reliability as we increase the block length.

A natural generalization of the RNN-based \o\ 
is to unroll the RNN cells. 
In Figure~\ref{noisy4} (Middle), we plot the BER as a function of the SNR, for 50 information bits and length 500 information bits (with noiseless feedback) when we unroll the RNN cells. We can see that the BER remains the same as we increase block lengths. 
This is not an entirely satisfying generalization because, typically, it is possible to design a code for which error rate decays faster as block length increases. For example, turbo codes have error rate decaying exponentially ($\log$ BER decades linearly) in the block length as shown in Figure~\ref{noisy4} (Right). This critically relies on the interleaver, which creates long range dependencies between information bits that are far apart in the block. Given that the neural encoder is a sequential code, there is no strong long range dependence. Each transmitted bit depends on only a few past information bits and their feedback (we refer to Section~\ref{sec5} for a detailed discussion). 

To resolve this problem, we propose a new concatenated code
which concatenates \o\ (as inner code) and turbo code as an outer code. 
 The outer code is not restricted to a turbo code, and we refer to Appendix~\ref{app:concat} for a detailed discussion.
In Figure~\ref{noisy4} (Right), we plot the BERs of the concatenated code, in channels with both noiseless and noisy feedback (of feedback SNR $10$dB), and turbo code, both at rate $1/9$ at (forward) SNR $-6.5$dB. From the figure, we see that
even with noisy feedback, BER drops almost exponentially ($\log$ BER drops linearly) as block length increases, and the slope is sharper than the one for turbo codes. 
 We also note that in this setting, C-L scheme suggests not using the feedback.

\section{Interpretation}\label{sec5}
Thus far we have used information theoretic insights in driving our deep learning designs. Here, we ask if the deep learning architectures we have learnt can provide an insight to the information theory of communications with feedback. We aim to understand the behavior of \o\ 
 (i.e., how coded bits are generated via RNN in Phase 2). We show that in the second phase, (a) the encoder focuses on refining information bits that were corrupted by large noise in the first phase; and (b) the coded bit depends on past as well as current information bits, i.e., coupling in the coding process. 

\begin{figure}[!ht]
\centering
\vspace{-0.8em}
\includegraphics[width=0.3\textwidth]{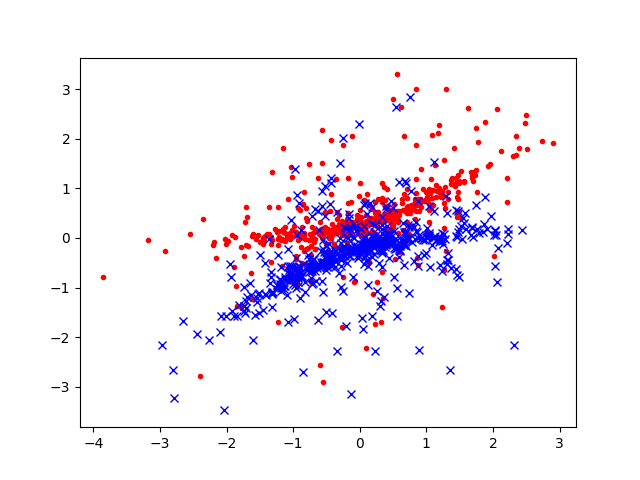}
\includegraphics[width=0.3\textwidth]{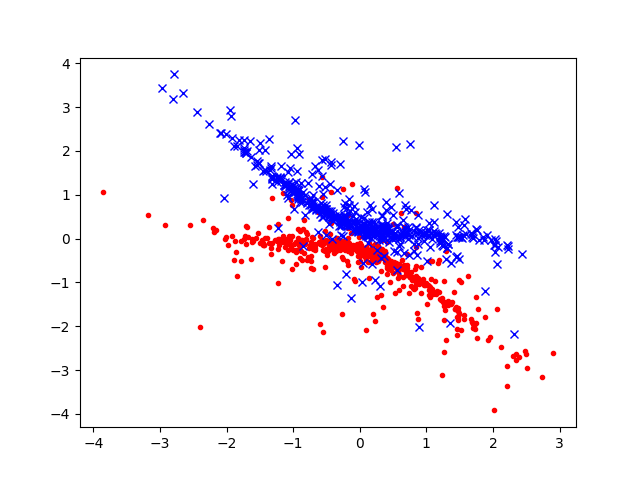}
	\put(-290,46){$c_{k,1}$}
	\put(-150,46){$c_{k,2}$}
	\put(-210,-3){$n_k$}
	\put(-85,-3){$n_k$}
	\caption{Noise in first phase $n_k$ vs. first parity bit $c_{k,1}$ (left) and second parity bit $c_{k,2}$ (right) under noiseless feedback channel and forward 
	AWGN channel of SNR 0dB. 
	Blue `\textcolor{blue}{x}' data points correspond to those samples conditioned on  
 	$b_k = 1$ and red `\textcolor{red}{o}' points correspond to those samples 
	conditioned on $b_k = 0$. }\label{noisepars}
\vspace{-0.3em}
\end{figure}

\bigskip\noindent
\textbf{Correcting  noise from previous phase.} 
The main motivation behind the proposed two-phase encoding scheme is 
to use the Phase 2 to clean the noise added in Phase 1. 
The encoder at Phase 2 knows how much noise was added in Phase 1 (exactly if noiseless feedback and approximately if noisy). 
Potentially, it could learn to send this information in the Phase 2, 
so that the decoder can refine the corrupted information bits sent in Phase 1.   
Interpreting the parity bits confirms this conjecture as shown in 
Figure~\ref{noisepars}. 
We show as a scatter plot multiple instances of the 
pairs of random variables $(n_k, c_{k,1})$ (left) and $(n_k, c_{k,2})$ (right), 
where $n_k$ denotes the noise added to the transmission of $b_k$ in the first phase. 
We are plotting 1,000 sample points: 20 samples for each $k$ and for $k\in\{1,\ldots,50\}$.  
This illustrates how the encoder has learned to send rectified linear unit (ReLU($x$)=$\max\{0,x\}$ ) functional of the noise $n_k$ to 
send the noise information while efficiently using the power.
Precisely, the dominant term in the parity bit can be 
closely approximated by $c_{k,1} \simeq -(2b_k-1) \times {\rm ReLU}(-n_k (2b_k-1))$, 
and $c_{k,2} \simeq (2b_k-1) \times {\rm ReLU}(-n_k (2b_k-1))$.

Consider the case when $b_k = 1$. 
If the noise added to bit $b_k$ in Phase 1 is positive, 
then the bit is likely to have been correctly decoded, and the parity chooses not to send any information about $n_k$.
The encoder generates coded bits close to zero (i.e., does not further refine $b_k$). 
Otherwise, the encoder generates coded bits proportional to the noise $n_k$, and hence uses more power to refine $b_k$.

 Ideally, for practical use, we want to use the same code for a broad range of varying SNR, 
 as we might be uncertain about the condition of the channel we are operating on.  
 This is particularly true for code for non-feedback channels. 
 For channels with output feedback, however, all known encoding schemes adapt to the channel.
 For instance, in S-K scheme, in order to achieve the optimal error rate, 
 it is critical to choose the optimal power allocation across transmission symbols
  depending on the (forward) channel SNR. 
Similarly, C-L scheme also requires pre-computation of optimal power allocation across transmission depending on the forward and feedback SNRs. 
In the case of AWGN channels with feedback, it is not even clear how one could meet the power constraints, if not adapting to the channel SNR. 
 
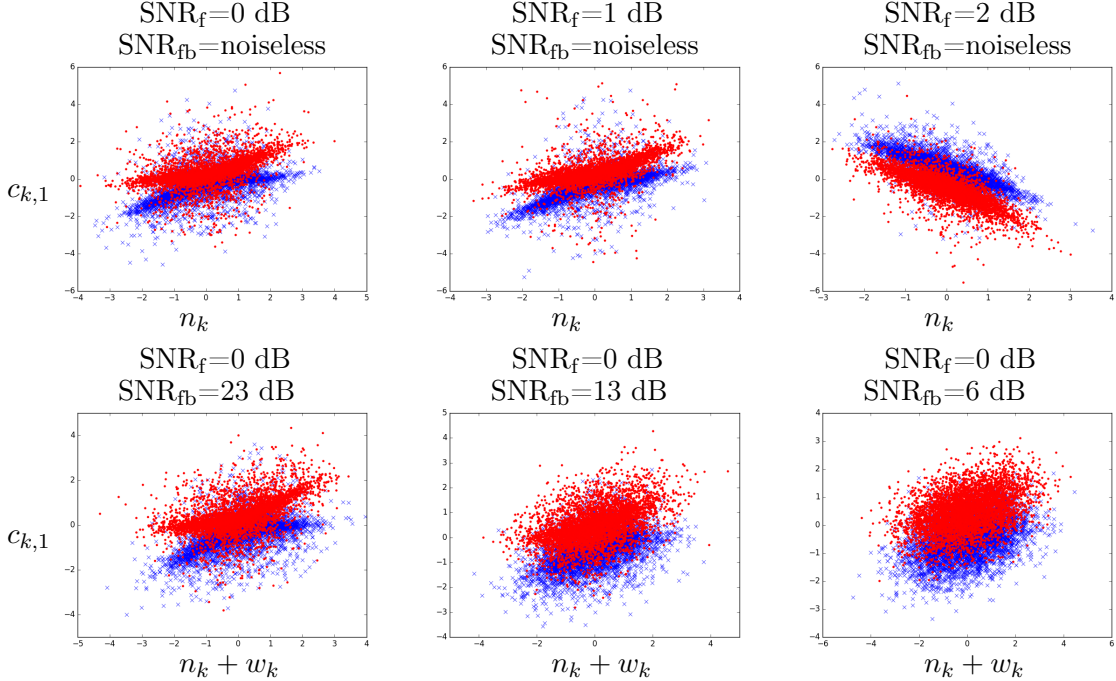
\begin{figure}[!ht]
	\centering
	\includegraphics[width=0.3\textwidth]{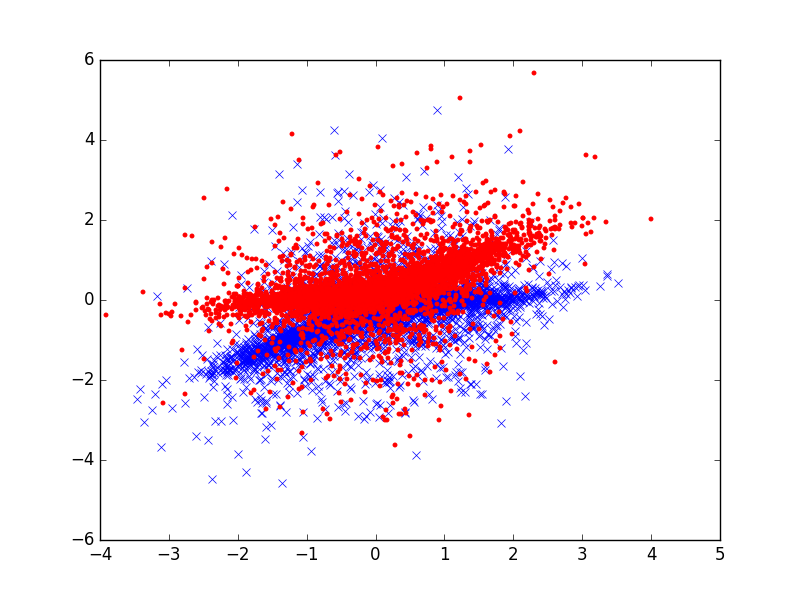}
	\put(-150,46){$c_{k,1}$}
	\put(-85,-3){$n_k$}
	\put(-100,112){${\rm SNR}_{\rm f}$=0 dB}
	\put(-108,100){${\rm SNR}_{\rm fb}$=noiseless}
	\includegraphics[width=0.3\textwidth]{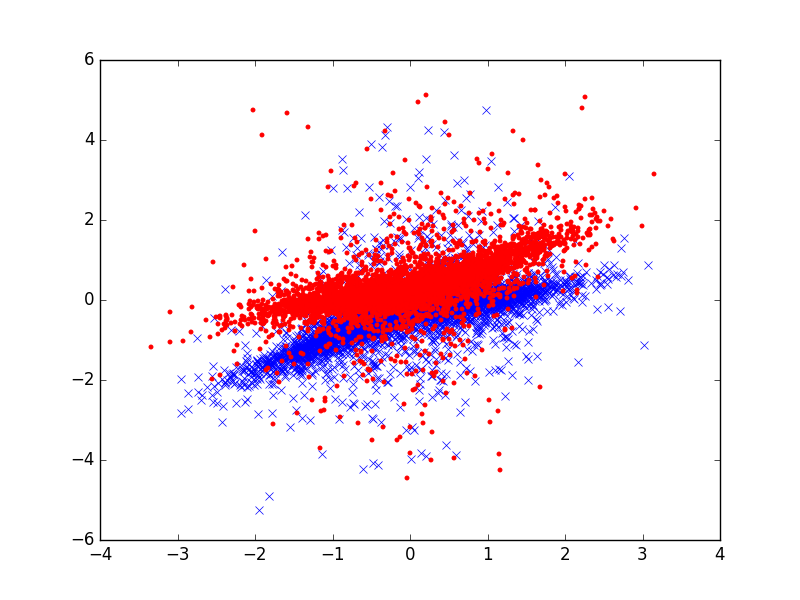}
	\put(-85,-3){$n_k$}
	\put(-100,112){${\rm SNR}_{\rm f}$=1 dB}
	\put(-108,100){${\rm SNR}_{\rm fb}$=noiseless}
	\includegraphics[width=0.3\textwidth]{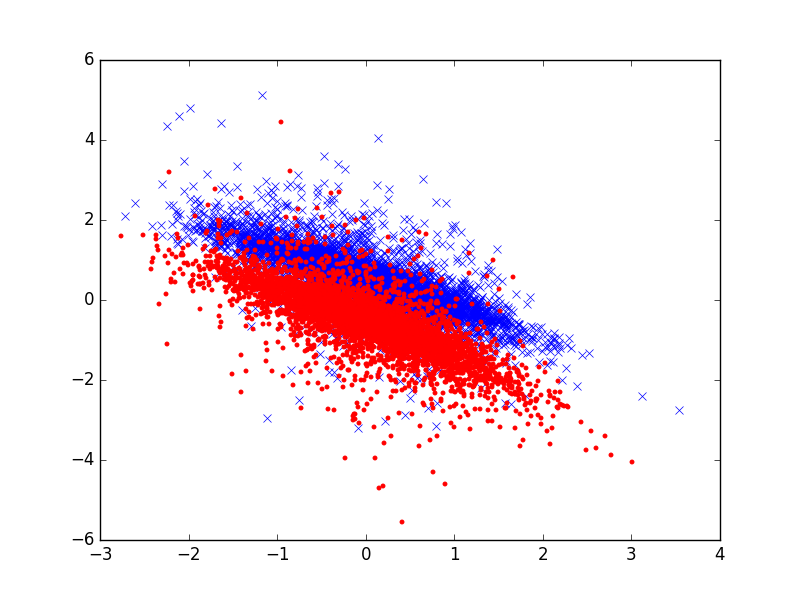}
	\put(-85,-3){$n_k$}
	\put(-100,112){${\rm SNR}_{\rm f}$=2 dB}
	\put(-108,100){${\rm SNR}_{\rm fb}$=noiseless}
	\vspace{0.2cm}
	\\
	
	\includegraphics[width=0.3\textwidth]{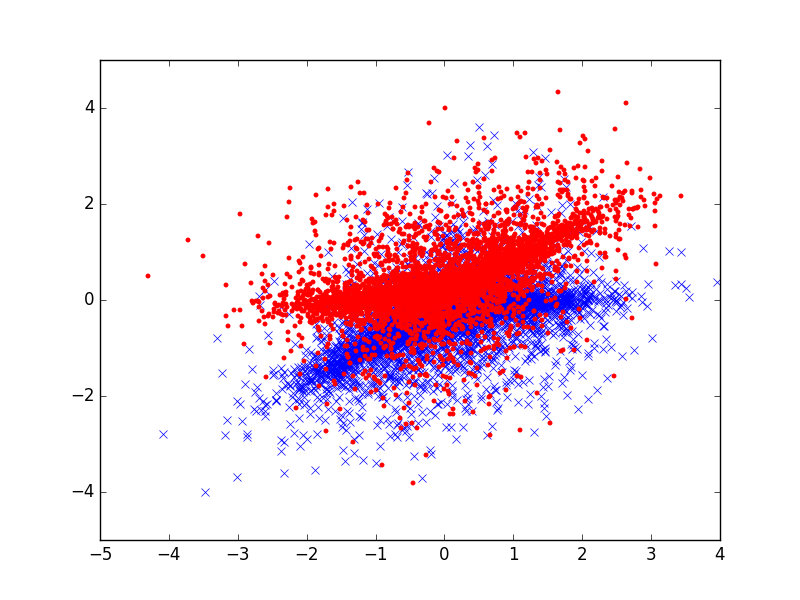}
	\put(-150,46){$c_{k,1}$}
	\put(-85,-3){$n_k+w_k$}
	\put(-100,112){${\rm SNR}_{\rm f}$=0 dB}
	\put(-108,100){${\rm SNR}_{\rm fb}$=23 dB}
	\includegraphics[width=0.3\textwidth]{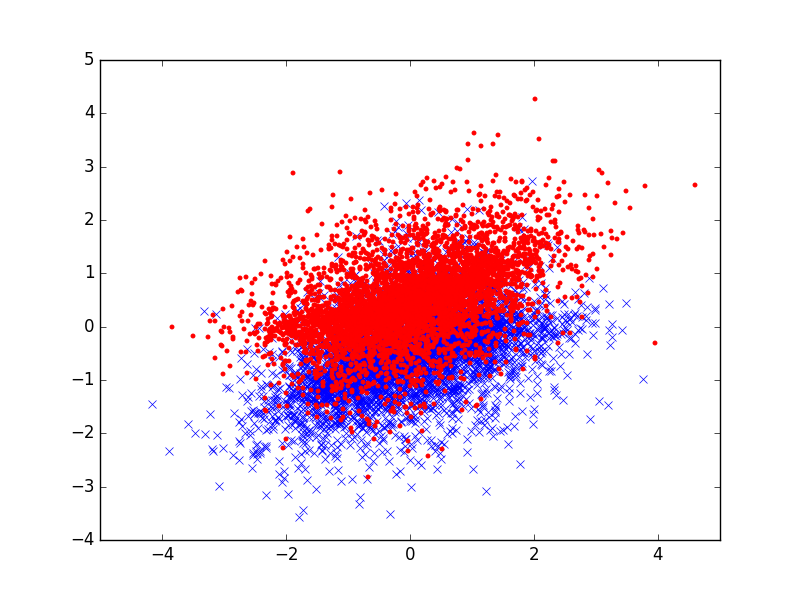}
	\put(-85,-3){$n_k+w_k$}
	\put(-100,112){${\rm SNR}_{\rm f}$=0 dB}
	\put(-108,100){${\rm SNR}_{\rm fb}$=13 dB}
	\includegraphics[width=0.3\textwidth]{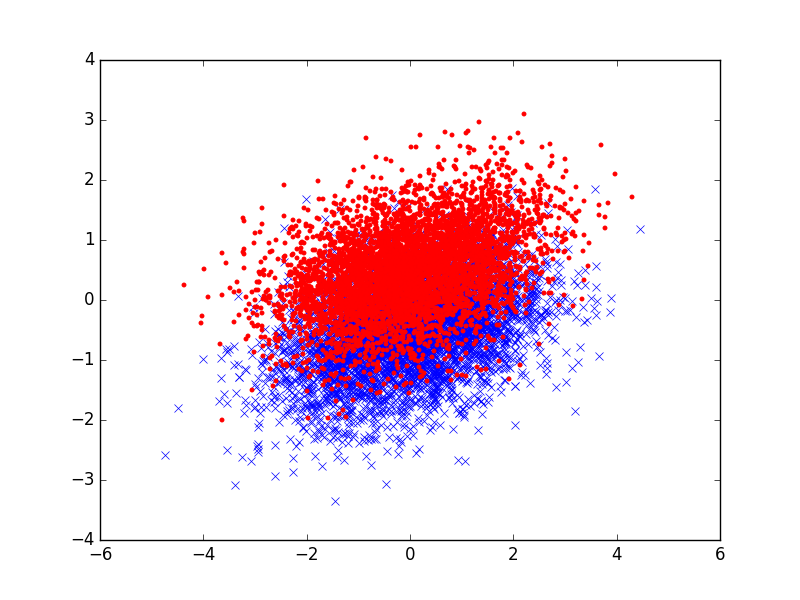}
	\put(-85,-3){$n_k+w_k$}
	\put(-100,112){${\rm SNR}_{\rm f}$=0 dB}
	\put(-108,100){${\rm SNR}_{\rm fb}$=6 dB}

	\caption{Noise in first phase $n_k$ vs. first parity bit $c_{k,1}$ as forward SNR increases for noiseless feedback (top) and sum of forward and feedback noises in first phase $n_k+w_k$ vs. first parity bit $c_{k,1}$ as feedback SNR decreases for fixed forward SNR 0dB (bottom). 	Blue `\textcolor{blue}{x}' data points correspond to those samples conditioned on  
 	$b_k = 1$ and red `\textcolor{red}{o}' points correspond to those samples 
	conditioned on $b_k = 0$. 
	}
	\label{fig:int_noise}
\end{figure}

 In Figure~\ref{fig:int_noise}, we show how the trained \o\ 
 has learned to adapt to the channel conditions. 
For various choices of 
forward channel ${\rm SNR}_{\rm f}$ and feedback channel ${\rm SNR}_{\rm fb}$, 
each scatter plot is  showing 5,000 sample points: 100 samples for each $k$ and for $k\in\{1,\ldots,50\}$.  
 On the top row, as forward signal power decreases,
the parity gradually changes from  $c_{k,1} \simeq -(2b_k-1) \times {\rm ReLU}(-n_k (2b_k-1))$ 
to $c_{k,1} \simeq (2b_k-1) -n_k $.
 On the bottom row, as feedback noise increases (i.e., feedback SNR decreases), 
 the parity gradually becomes less correlated with the sum of forward and feedback noises $n_k + w_k$. Note that under noisy feedback, $n_k$ is not available to the encoder, but $n_k + w_k$ is what is available to the encoder ($n_k + w_k = \tilde{y}_k - c_k$).  




 \bigskip\noindent
\textbf{Coupling.} 
A natural question is whether our feedback code is exploiting the memory of RNN and coding information bits jointly. 
To answer this question, we look at the correlation between information bits and the coded bits. 
If the memory of RNN were not used, we would expect the coded bits $(c_{k,1},c_{k,2})$ to depend only on $b_k$. 
We find that $\E[c_{k,1}b_{k}] = -0.42, \E[c_{k,1}b_{k-1}] = -0.24, \E[c_{k,1}b_{k-2}] = -0.1, \E[c_{k,1}b_{k-3}] = -0.05$, and  $\E[c_{k,2}b_{k}] = 0.57, \E[c_{k,2}b_{k-1}] = -0.11, \E[c_{k,2}b_{k-2}] = -0.05, \E[c_{k,2}b_{k-3}]$ $ = -0.02$ (for the encoder for forward SNR 0dB and noiseless feedback). This result implies that the RNN encoder does make use of the memory, of length two to three. 

Overall, our analysis suggests that \o\ 
 exploits memory and selectively enhances bits that were subject to larger noise - properties reminiscent of any good code. We also observe that the relationship between the transmitted bit and previous feedback demonstrates a non-linear relationship as expected. Thus our code has features requisite of a strong feedback code. Furthermore, improvements can be obtained if instead of transmitting two coded symbols per bit during Phase 2, an attention-type mechanism can be used to zoom in on bits that were prone to high noise in Phase 1. These insights suggest the following generic feedback code: it is a sequential code with long cumulative memory but the importance of a given bit in the memory is {\em dynamically} weighted based on the feedback. 



%
%
%
%





\section{System and implementation issues}\label{sec6}
We began with the idealized Shannon model of feedback and have progressively considered practical variants (delay, noise and active feedback). In this section we extend this progression by studying design decisions in real-world implementations of \o\, (our neural-network feedback-enabled codes). We do this in the context of cellular wireless systems, with specific relevance to the upcoming 5G LTE standard.

LTE cellular  standards prescribe separate uplink and downlink transmissions (usually in frequency division duplex mode). Further, these transmissions are scheduled in a centralized manner by the base station associated with the cell. In many scenarios, the traffic flowing across uplink and downlink could be asymmetric (example: more ``downloads'' than ``uploads'' leads to higher downlink traffic than the combined uplink ones). In such cases, there could be more channel resources  in the uplink than the traffic demand.  Given the sharp inflexible division among uplink and downlink, these resources go unused. We propose to link, {\em opportunistically}, unused resources in one direction to aid the reliability of transmission in the opposite direction -- this is done via using the feedback codes developed in this paper. Note that the availability of such unused channel resources is known in advance to the base station which makes scheduling decisions on {\em both} directions of uplink and downlink -- thus such a synchronized cross uplink-downlink scheduling is readily possible.

\begin{figure}[!ht]
\centering
\includegraphics[width=0.42\textwidth]{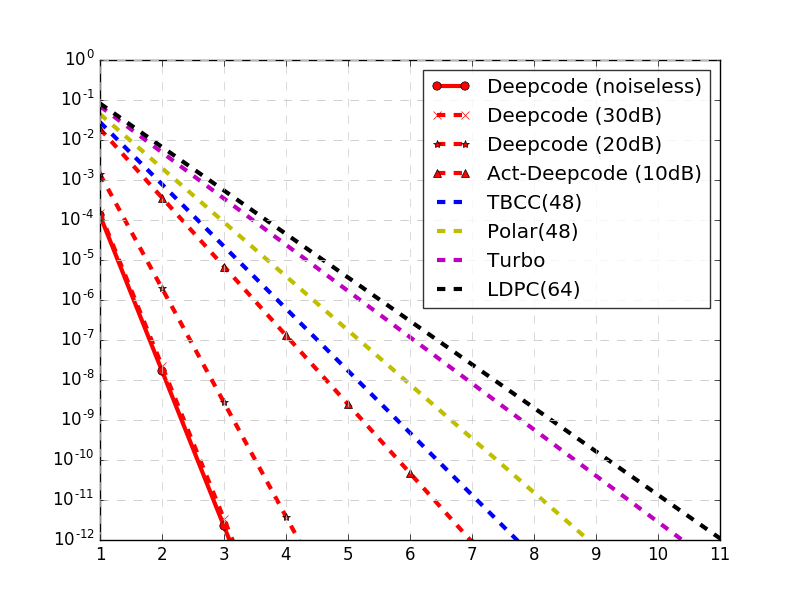}
	\put(-207,139){BLER}
	\put(-152,-6){Number of transmissions}
	\caption{BLER as a function of number of transmissions for a rate 1/3 code with 50 information bits where forward SNR is 0dB. \O\ allows fewer transmissions than feedforward codes to achieve the target BLER.}\label{arq_0}
\end{figure}

The availability of the feedback traffic channel enables the usage of the codes designed in this paper -- leading to much stronger reliability than the feedforward codes alone. Combined with automatic repeat request (ARQ), this leads to fewer retransmissions and smaller average transmission time than the traditional scheme of feedforward codes combined with ARQ would achieve. 
In order to numerically evaluate the expected benefits of such a system design, 
in Figure~\ref{arq_0}, we plot BLER as a function of number of (re)transmissions for \o\ under noiseless and noisy feedback and feedforward codes (for a rate $1/3$ code with 50 information bits). From this figure, we can see that combining \o\  with ARQ allows fewer block transmissions to achieve the target BLER compared to the state-of-the-art codes. 
The performance of \o\  depends on the quality of the feedback channel. As feedback channel becomes less noisy, \o\ requires fewer retransmissions.   
We note that in measuring the BLER of neural code under noisy feedback (10dB), we used a variant of \o, shown as Act-Deepcode, which allows an active feedback of rate $3/4$; in Phase 1, 
the decoder sends back RNN encoded bits at rate 1/2. Phase 2 works as in \o. Hence, for a rate 1/3 code with 50 information bits, the decoder makes 200 usages of the feedback channel (204 with zero padding). Improving further the performance of (active) \o\ at realistic feedback SNRs (such as 10dB or lower) is an important open problem. The improvements could come from architectural or learning methodology innovations or a combination of both.


We propose using \o\  when the feedback SNR is high. 
Practically, a user may not always have a high SNR feedback channel, but when there are multiple users, it is possible that some of the users have high SNR feedback channels. 
For example, in scenarios where a base station communicates with multiple users, we propose scheduling users based on their feedback as well as forward channel qualities, 
utilizing multiuser diversity. 
In Internet-of-Things (IoT) applications, feedback channel SNR can be much higher than forward SNR; e.g., a small device with limited power communicates a message to the router connected to the power source.

\section{Conclusion}\label{conclusion}
In this paper we have shown that appropriately designed and trained RNN codes (encoder and decoder){, which we call \o, }
 outperform the state-of-the-art codes by a significant margin on the challenging problem of communicating over AWGN channels with noisy output feedback, both on the theoretical model and with practical considerations taken into account. 
%
%
 By concatenating \o\ 
 with a traditional outer code, the BER curve drops significantly with increasing block lengths, allowing generalizations of the learned neural network architectures. 
 The encoding and decoding capabilities of the RNN architectures suggest that new codes could be found in other open problems in information theory (e.g., network settings), where practical codes are sorely missing. 
Our work suggests several immediate avenues continue the research program  begun by this paper, solutions to which will have  significant practical impacts. 

\bigskip
\noindent
{\bf Learning to take advantage of the block lengths.} 
The first one is an interesting new challenge for machine learning, that has not been posed before to the best of our knowledge. 
We proposed concatenation in Section \ref{sec4} to achieve the block-length gain. 
 By concatenating \o\ 
  with a traditional inner code, 
 the BER curve drops significantly with increasing block lengths, 
 allowing generalizations of the learned neural network architectures. 
 However, concatenation comes at the cost of reduced rate. 

A more natural way to achieve the block-length gain is to 
incorporate structures of the modern codes, in particular turbo codes. 
Turbo codes use inter-leavers to introduce long range dependency on top of standard convolutional codes, 
achieving error rate that exponentially decays in block-lengths as desired.  
In the encoder, we can easily include an inter-leaver with two neural encoders we proposed. 
However, the major challenge is in decoding. 
Turbo decoder critically relies on BCJR decoder's accurate estimate of the posterior probability of each information bit. 
This is in turn fed into the next phase of turbo decoder, which refines the likelihood iteratively. 
For the proposed neural feedback code, there exists no decoder that can output accurate posterior likelihood. 
Further, there exists no decoder that can take as part of the input the side information from the previous phase on the prior likelihood of the information bit. 

This poses an interesting challenge for deep learning. 
In machine learning terminology, 
we consider a supervised learning setting with binary classification, 
but instead of asking for accurate classification (as measured by average loss $(1/n)\sum_{i=1}^n \ell( f(X_i), Y_i) $) 
of the example $X_i$ to predict the label $Y_i$ on some loss function $\ell(\cdot,\cdot)$, 
we ask for accurate likelihood.
We need an accurate estimate of ${\mathbb P}(Y_i=1|X_i)$, which could be measured by 
$(1/n)\sum_{i=1}^n \ell( f(X_i), {\mathbb P}(Y_i=1|X_i)) $. 
This is a much more challenging task than traditional supervised learning, which is necessitated because the supervision we desire, i.e.~${\mathbb P}(Y_i=1|X_i)$, 
is not readily available. 
On the other hand, as an infinite amount of training data can be readily generated under the communication scenario, 
there is a hope that with the right algorithm, the posterior probability can be predicted accurately. 
Such a solution will be a significant contribution not only to communication algorithms, but also to the broader machine learning community. 

\bigskip
\noindent
{\bf Interpreting \o. }
The second challenge is in using the lessons learned from the trained \o\ to contribute back to the communication theory. 
We identified in Section \ref{sec5} 
some parts of the parity symbols of the trained \o. 
However, how \o\ 
 is able to exploit the feedback symbols remains mysterious, despite our efforts to interpret the trained neural network. 
It is an interesting challenge to disentangle  the neural encoder, 
and provide a guideline for designing simple feedback encoders that enjoy some of the benefits of the complex neural encoder. 
Manually designing such simple encoders without training 
can provide a new family of feedback encoders that are simple enough to 
be mathematically analyzed.

\bigskip 
\noindent
{\bf Rate beyond 1/3.} 
The third challenge is to generalize \o\ 
to rates beyond 1/3. Our neural code structure can be immediately generalized to rates $1/r$ for $r = 2, 3, 4, \cdots$. For example, we have preliminary results showing that a rate-$1/2$ RNN based feedback code beats the state-of-the-art codes for short block lengths (e.g., 64) under low SNRs (e.g., below 2dB). Extensive experiments and simulations over various rates and comparison to state-of-the-art codes are yet to be explored. 
On the other hand, generalization to rates higher than $1/2$ requires a new architecture of encoders. In this direction, we propose two potential approaches. One is to use a higher-order modulation (e.g., pulse amplitude modulation) and generate parity bits for super symbols which are functions of multiple information bits. The other is to use puncturing, a widely used technique to design high rate codes from low rate codes (e.g., convolutional codes); the encoder first generates a low rate code and then throws away some of the coded bits and sends only a fraction of the coded bits. Generalization to higher rate codes via these two approaches is of great practical interest.


%
%
%
%
%
%
%



\section*{Acknowledgement}
We thank  Shrinivas Kudekar and Saurabh Tavildar for helpful discussions and providing references to the state-of-the-art feedforward codes.  We thank Dina Katabi 
for a detailed discussion that prompted the work on system implementation. 
\medskip

\small

\bibliographystyle{IEEEtran}
\bibliography{arxiv_nips2018.bib}

\newpage
\appendix
\section*{Appendix}
\label{appendix}
\section{State-of-the art codes used in comparison}\label{5G_doc}
In this section, we provide details on how to compute the BER and BLER of state-of-the art feedforward codes. 
LTE turbo code used in the simulation uses trellis-([13, 15], 13) convolutional code (octal notation) as a component code, and uses quadratic permutation polynomial (QPP) interleaver. 
Decoding is done by 8 iterations of Belief Propagation (BP) decoder that uses a posteriori probability (APP) decoder as the constituent decoder.  
Tail-bitting convolutional codes (TBCC) used in the simulation has a constraint length 7 and trellis ([123,135,157]) (in octal notation), and uses Viterbi decoder. 
Polar code used in the simulation uses success cancellation list decoding (SCL) with list size 8. 
LDPC code used in the simulation (Rate 1/3, maps 64 bits to a length-196 codeword with sub-matrix dimension 16) uses the parity check matrix shown below, and layered offset min-sum decoder is used with offset parameter 0.22 and (max) iteration 25. 
\[
\begin{bmatrix} 
  10 & 11 &  2  & 3  & 0   & -1 & -1 & -1 & -1 & -1 & -1 & -1 \\
  -1 & 15  &  9 & 9  & 14  &  0 & -1 & -1 & -1 & -1 & -1 & -1\\
   6 & -1  &  5  & 13 & -1  & 11 &  0 & -1 & -1 & -1& -1 & -1\\
  -1 & 5  & -1  &  8  & 12  & -1 &  6 & 0  & -1 & -1 & -1 & -1\\
  -1 & 11 & -1  & -1 &  1  & -1  & -1 & 11 &  0 & -1 & -1 & -1\\
  -1 &  2 & -1  & -1  & 14  & 12& -1 & 7  & -1 &  0 & -1 & -1\\
  -1 & 15 & 10 & -1 & -1  & -1  & -1 & -1 & 11 & -1 & 0 & -1\\
  -1 & -1 & -1  & 7   &  -1 & 11 & -1 &  3  & -1 & -1 & -1&  0\\
\end{bmatrix}
\]

\section{Implementation details}
\label{sec:implement}

In this section, we provide implementation details on the neural encoders and decoders, introduced in Section~\ref{sec3}, for the AWGN channels with feedback. 

\subsection{Illustration on Scheme A. RNN feedback encoder/decoder (RNN (linear) and RNN (tanh)).}\label{schemeA}
\begin{figure}[!ht] 
\centering
\includegraphics[align=c, width=0.6\textwidth]{figs/enc1.png}\ \ \ 
\includegraphics[align=c, width=0.35\textwidth]{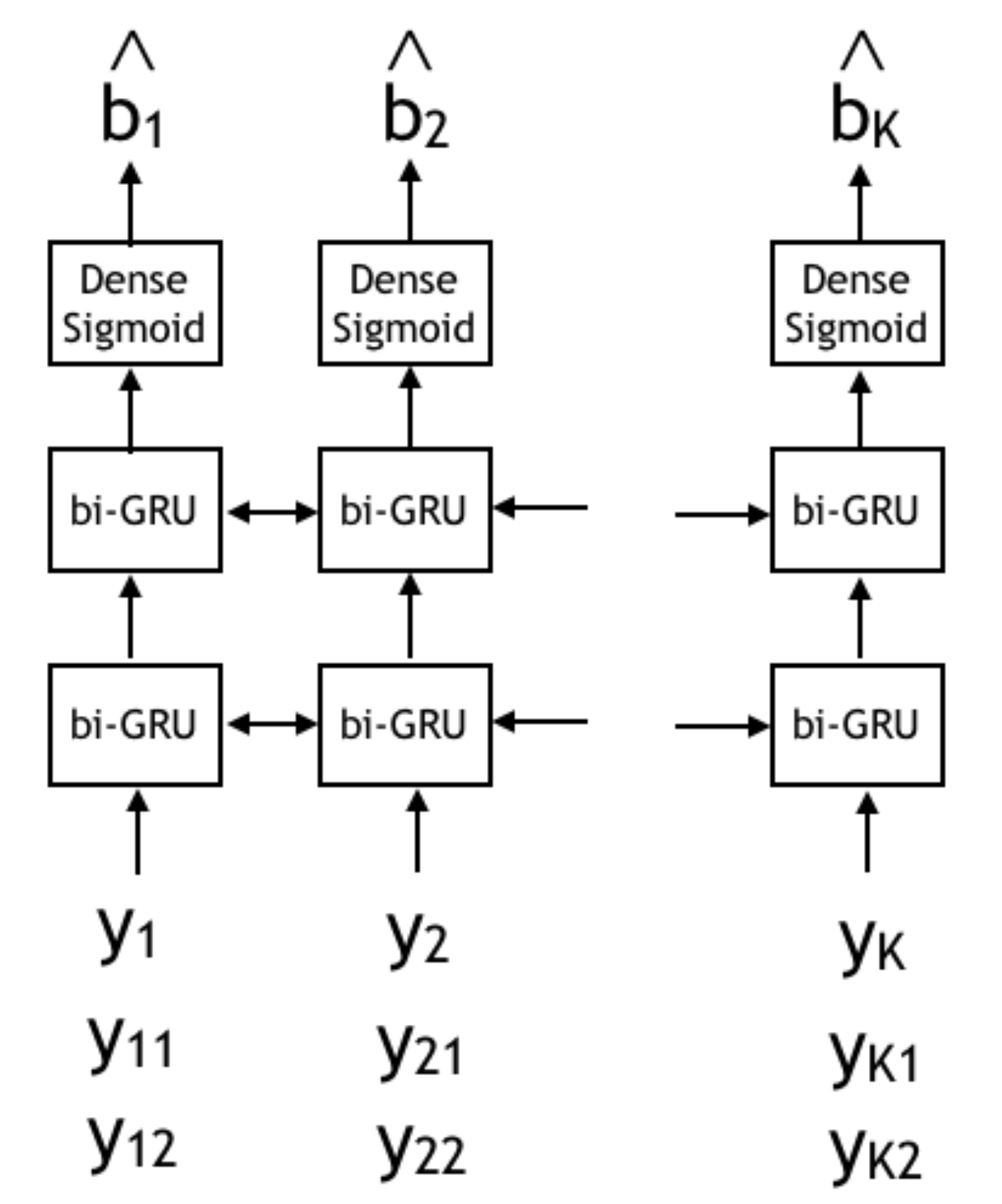}
\caption{RNN feedback encoder (left) and decoder (right)}\label{encdecdiag}
\end{figure}

The details of neural encoder and decoder architectures for RNN feedback code are illustrated in Table~\ref{feedback_enc} and Figure~\ref{encdecdiag}. The architectures for RNN (tanh) and RNN (linear) feedback codes are equivalent except for the activation function in RNN; RNN (tanh) encoder uses a tanh activation while RNN (linear) encoder uses a linear activation (for both the recurrent and output activation). 



\begin{table}[!ht]
	\caption{Architecture of RNN feedback encoder (left) and decoder (right) for AWGN channels with noisy feedback.}
	\label{feedback_enc}
\centering
\begin{tabular}{ |c |c|}
\hline
 Layer & Output dimension\\
\hline
Input & (K, 4)\\
RNN (linear or tanh)& (K, 50)\\
Dense (sigmoid) & (K, 2)\\
Normalization & (K, 2)\\
 \hline
\end{tabular}\ \ \ \ \ 
\begin{tabular}{ |c |c|}
\hline
 Layer & Output dimension\\
\hline
Input & (K, 3)\\
bi-GRU & (K, 100)\\
Batch Normalization & (K, 100)\\
bi-GRU & (K, 100)\\
Batch Normalization & (K, 100)\\
Dense (sigmoid) & (K, 1)\\
 \hline
\end{tabular}
\end{table}

\subsection{Illustration on Scheme B. RNN feedback code with zero padding (RNN (tanh) + ZP).}\label{schemeB}
\begin{figure}[!ht] 
\centering
\includegraphics[width=0.5\textwidth]{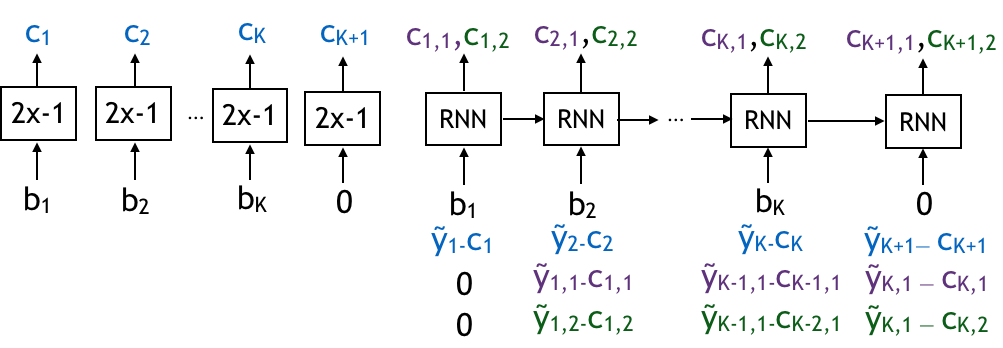} 
\caption{Encoder for scheme B.}\label{enc2}
\end{figure}

\begin{figure}[!ht] 
\centering
\includegraphics[width=0.4\textwidth]{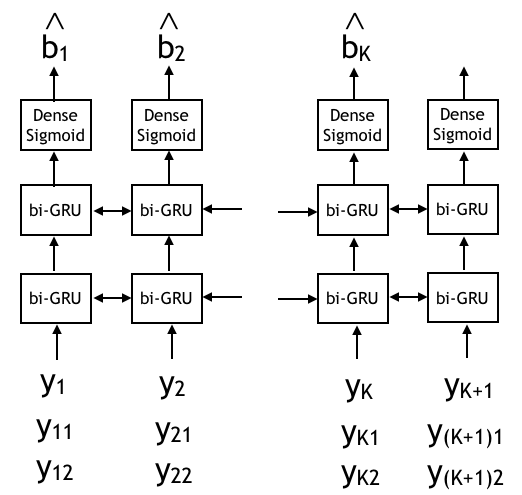}
\caption{Decoder for schemes B,C,D.}\label{decoder2}
\end{figure}

The encoder and decoder structures with zero padding are shown in Figure~\ref{enc2} and Figure~\ref{decoder2}, respectively. We maintain the encoder and decoder architecture same as Scheme A (RNN (tanh)) and simply replace the input information bits by information bits padded by a zero; hence, we use $K+1$ RNN cells in Phase 2 instead of $K$.
In training, we use back-propagation with binary cross entropy loss as we did in Scheme A. A slight modification is that  
we measure binary crossentropy loss on the information bits of length $K$ only (i.e., ignore the loss on the last bit which corresponds to a zero padding).

\subsection{Illustration on Scheme C. RNN feedback code with power allocation\\ (RNN(tanh) + ZP + W).}\label{schemeC}
\begin{figure}[!ht] 
\centering
\includegraphics[width=0.6\textwidth]{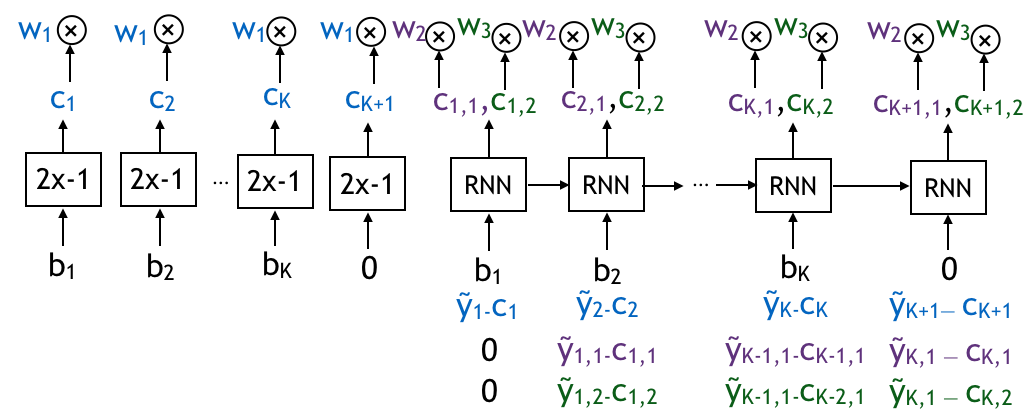} 
\caption{Encoder C.}\label{enc3}
\end{figure}

The encoder and decoder architectures for scheme C are shown in Figure~\ref{enc3} and Figure~\ref{decoder2}, respectively. 
Specifically, we introduce three trainable weights $(w_0,w_1,w_2)$ and let
$\E[c_k^2] = w_0^2, \E[c_{k,1}^2] = w_1^2, \E[c_{k,2}^2] = w_2^2$ for all $k \in \{1,\cdots,K\}$ where $w_0^2 + w_1^2 + w_2^2 = 3$ (c.f. in Encoder B, we let $\E[c_k^2] = \E[c_{k,1}^2] = \E[c_{k,2}^2] = 1$).
 In training, we initialize $w_i$s by $1$ and train the encoder and decoder jointly as we trained Schemes A and B.
The trained weights are $(w_1,w_2,w_3) = (1.13,0.90, 0.96)$ (trained at -1dB). This implies that the encoder uses more power in Phase I, to transmit (raw) information bits. In Phase II, the encoder uses more power on the second parity bits than in the first parity bits.

\subsection{Scheme D. RNN feedback code with bit power allocation (RNN(tanh) + ZP + W + A).}\label{schemeD}
\begin{figure}[!ht] 
\centering
\includegraphics[width=0.6\textwidth]{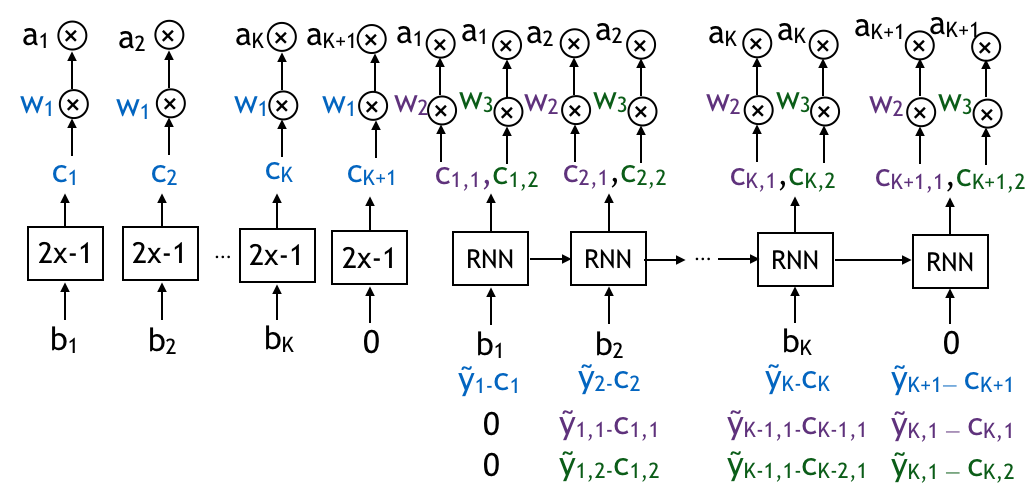}
\caption{Encoder D.}\label{enc4}
\end{figure}

The encoder and decoder architectures for scheme D are shown in Figure~\ref{enc4} and Figure~\ref{decoder2}, respectively. 
We introduce trainable weights $a_1, a_2, \cdots, a_K,a_{K+1}$ for power allocation in each transmission. To the full generality, we can train all these $K+1$ weights. 
However, we let $a_5,\cdots, a_{K-5} = 1$ and only train first 4 weights and the last 5 weights, $a_1,a_2,a_3,a_4$ and $a_{K-4},a_{K-3},a_{K-2},a_{K-1},a_{K},a_{K+1}$, for two reasons. Firstly, this way we can generalize the encoder to longer block lengths by maintaining the weights for first four and last five weights and fixing the rest of weights as 1s, no matter how many rest weights we have. For example, if we test our code for length 1000 information bits, we can let $a_5, \cdots, c_{996} = 1$. Secondly, the BERs of middle bits do not depend much on the bit position; hence, power control is not needed as much as in the first and last few bits.  

In training scheme D, we initialize the encoder and decoder as the ones in Scheme C, and then 
additionally train the weight vectors $\textbf{a}$ on top of the trained model, while allowing all weights in the encoder and decoder to change as well. After training, we see that 
the trained weights are $(a_1,a_2,a_3,a_4) = (0.87, 0.93, 0.96, 0.98)$
 and $(a_{K-4},a_{K-3},a_{K-2},a_{K-1},a_{K},a_{K+1}) = (1.009, 1.013, 1.056, 1.199, 0.935)$ (for $-1dB$ trained model). As we expected, the trained weights in the later bits are larger. Also, the weight at the $K+1$th bit position is small because last bit is always zero and does not convey any information. On the other hand, trained weights in the beginning positions are small because these bits are naturally more robust to noise due to the sequential structure in generation of parity bits in Phase 2. 

\subsection{Feedback with delay and coding}\label{sec:delayedfb}

Practical feedback typically is delayed for a random time, thus the encoder cannot use immediate feedback to encode. The feedback is randomly delayed up to block length $K$, we are restricted not to use feedback till $K$ bits are transmitted. Coding in both forward and feedback channel under noisy feedback will strengthen the reliability of communication.

\begin{figure}[!ht] 
\centering
\includegraphics[width=0.5\textwidth]{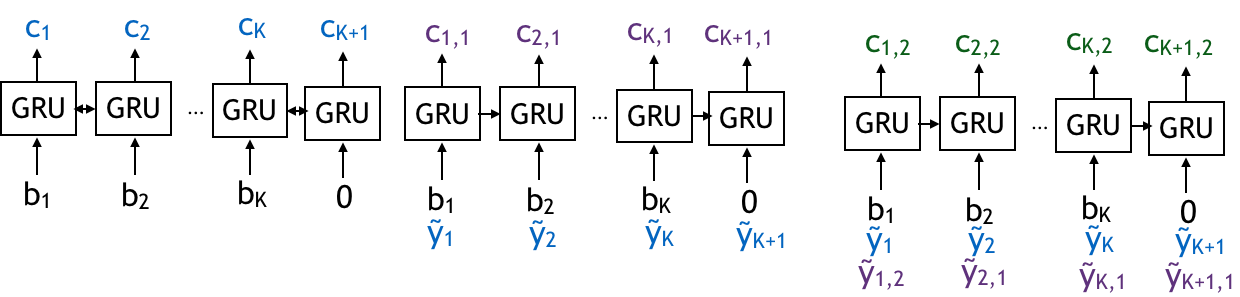}
\caption{Encoder for delayed feedback}\label{delay1}
\end{figure}

We propose an active and delayed feedback scheme to overcome noisy feedback and delaying effect; the 1/3 code rate encoder is shown in Figure~\ref{delay1}.  In the first phase, the $K$ information bits can be encoded by Bi-GRU, while the feedback is delayed and can only be used in the next phase. The second and third phases use uni-directional GRU to encode with $K$-delayed feedback, which means at index $m$ of phase 2, the encoder can only use the feedback before index $m$ of phase 1. Receiver side encodes the feedback by unidirectional GRU and sends through the delayed feedback channel back to the transmitter. The decoder is a Bi-GRU which waits to decode until all information bits are received.

\begin{figure}[!ht] 
\centering
\includegraphics[width=0.5\textwidth]{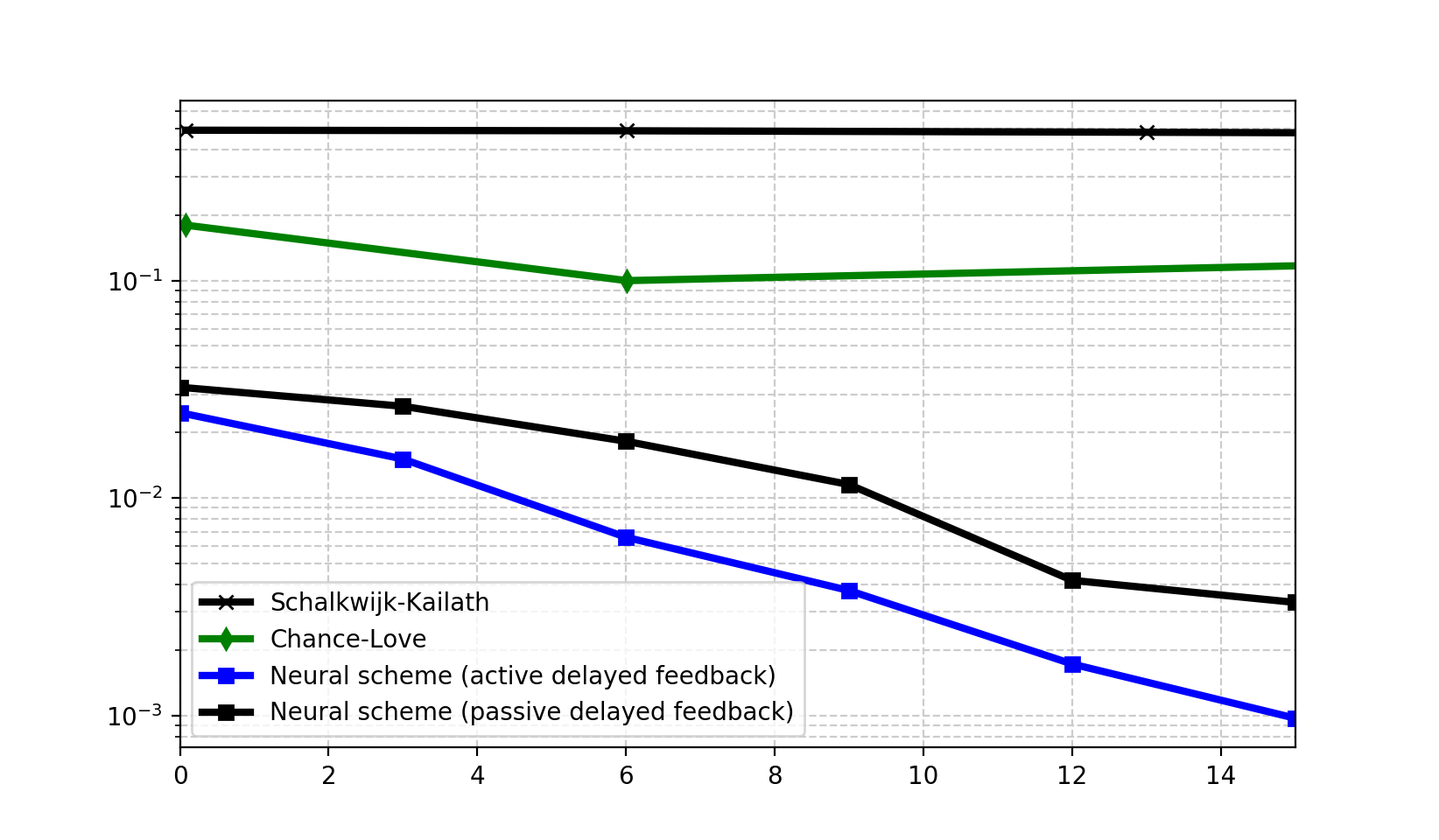}
\includegraphics[width=0.4\textwidth]{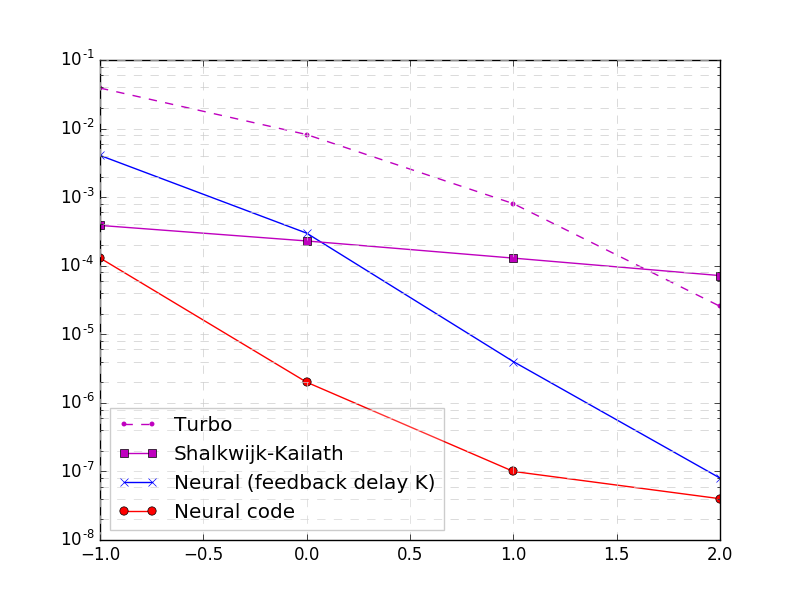}
\caption{Neural schemes for delay with feedback under noisy (left) and noiseless (right) feedback}\label{delay2}
\end{figure}

We can see from Figure ~\ref{delay2} (Right) that passive feedback under delayed feedback still has better performance compared to the turbo code, and beats S-K code under high SNR regimes. The delaying effect is enabled  via our RNN feedback coding scheme. The gain is from: (1) adding an additional phase, which gives the RNN more fault tolerance compared to 2-phase coding; (2) training the RNN to decode with delayed feedback.

Figure ~\ref{delay2} (Left) shows the performance under noisy feedback. The forward channel is under AWGN 0dB, while the x-axis shows the feedback SNR. The C-L and S-K codes fail to decode under noisy feedback channel. Passive feedback code achieves better performance comparing to C-L and S-K code, while active feedback code outperforms passive feedback code. The performance gain is from: (1) the coding gain of active feedback, which gives the encoder RNN better robust representation of feedback code; (2) as the feedback is noisy, delayed coding actually averages the noise, which leads to better performance.

\noindent  \textbf{Literature on coded feedback}\label{relv}
In~\cite{Shayevitz_active}, the authors show that active feedback can improve the reliability under noisy feedback if the feedback SNR is sufficiently larger than the forward SNR. Their coding scheme assumes that the encoder and decoder share a common random i.i.d.\ sequence (of length equals to the coded block length), mutually independent of the noise sequences and the message. We do not assume such a common randomness for the encoder and decoder; hence, we cannot directly compare our scheme with theirs. 

\section{Concatenation of neural code with existing codes}
\label{app:concat}
Concatenated codes are constructed from two or more codes, originally proposed by Forney~\cite{Forney_Concat}. We concatenate forward error correcting codes (that do not use a feedback) with our neural code that makes use of feedback. Encoding is performed in two steps; we first map information bits into a turbo code, and then encode the turbo code via an encoder for channels with feedback. Decoding is also performed in two steps. In the first step, the decoder recovers the estimates of turbo codes. In the second step, the decoder recovers information bits based on the estimates of turbo codes. For the experiment in Section~\ref{sec4}, for which results are shown in Figure~\ref{noisy4} (Right), 
we use the rate 1/3 LTE turbo code as an outer code; LTE turbo code uses ([13, 15], 13) convolutional code (octal notation) as a component code. We compare the performance of the concatenated code with a rate 1/9 turbo code, which uses ([13,17,16,15,11],13) convolutional code as a component code (introduced in~\cite{turbo_rate1_9}). Besides turbo codes, any existing codes (e.g., LDPC, polar, convolutional codes) can be used as an outer code. We also note that C-L scheme is based on the concatenation idea~\cite{Chance--Love2011}.

\section{Existing codes: C-L and S-K schemes}
\label{sec:CLSK}


In this section, we provide an illustration of two baseline schemes, C-L scheme and S-K scheme, and the connection between these schemes and our neural codes.

A simple scheme is to linearly encode each information bit separately using feedback. For each bit $b_k$, the encoder generates three coded bits $(c_{k1},c_{k2},c_{k3})$. This is the Chance-Love scheme proposed in~\cite{Chance--Love2011}. One of the contributions of~\cite{Chance--Love2011} is to empirically find the optimal weights for the linear functions (there is no closed-form solution). Another contribution is that they propose concatenating their code with an existing forward error correction code such as turbo codes, i.e., instead of mapping the information bits $\textbf{b}$ directly to the codeword $\textbf{c}$, the encoder maps $\textbf{b}$ to a turbo code $\textbf{d}$ and then maps the turbo code $\textbf{d}$ to a codeword $\textbf{c}$. 
\begin{figure}[!ht] 
\centering
\includegraphics[width=0.25\textwidth]{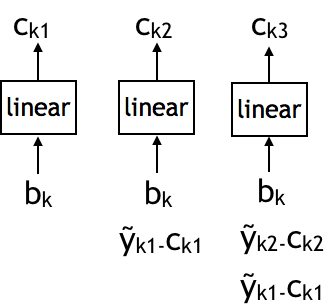}\ \ \ \
\caption{Illustration of encoding of $k$-th bit for a rate-1/3 linear encoder in Chance-Love scheme}\label{bit-by-bit}
\end{figure}



\begin{figure}[!ht] 
\centering
\includegraphics[width=0.25\textwidth]{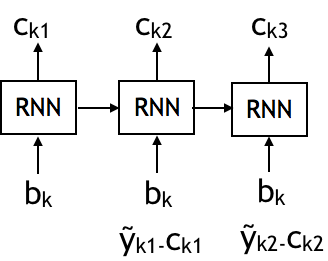}
\caption{Encoding of $k$-th bit for a rate-1/3 RNN encoder}\label{simple111}
\end{figure}

Can we start with a neural architecture that includes the C-L as a special case and improve upon it? Due to the sequential nature of feedback encoder, recurrent neural network (RNN) architectures are natural candidates. A simple neural architecture that includes the C-L scheme as a special case is illustrated in Figure~\ref{simple111}. We consider various versions of RNN encoders --RNN with linear activation functions, and nonlinear RNN, GRU, LSTM. 
We train the encoder and decoder jointly.
For all architectures, we use 
 50 hidden units.
The BERs of trained networks are shown in Table~\ref{bit-by-bit-ber}.
We can see that the BER of nonlinear RNN is smaller than that of a linear feedback scheme with weights optimized.

\begin{table}[!ht]
\centering
\begin{tabular}{ |c |c|c|c|}
\hline
 Scheme & BER at $1dB$ ($\sigma_F^2=0.01$)\\
\hline
Bit-by-bit linear (Shalkwijk-Kailath) &  0.0023\\
Bit-by-bit linear (Chance-Love) &7.83e-04 \\
Bit-by-bit linear RNN & 0.0046\\
Bit-by-bit RNN & \textbf{1.56e-04}\\
Bit-by-bit GRU & 1.58e-04\\
Bit-by-bit LSTM & {1.88e-04}\\
 \hline
\end{tabular}
\caption{BER of other RNN architectures. Rate 1/3}\label{bit-by-bit-ber}
\end{table}

Although RNN has the capability to represent any linear bit-by-bit linear encoder/decoder, we can see that the training is highly nontrivial, and for linear RNN, the neural network converges to a local optimum. 
On the other hand, for nonlinear RNNs, the trained encoder performs better than the weight-optimized linear scheme. 

\begin{figure}[!ht] 
\centering
\includegraphics[width=0.3\textwidth]{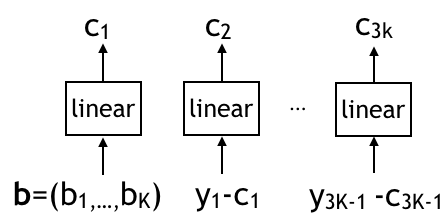}
\caption{Illustration of S-K encoder}\label{KSenc}
\end{figure}

From coding theory, we know that the bit error rate should go down as block length gets longer. If we use  bit-by-bit encoding, the improvement can never be realized because BER remains the same now matter how long the block is. In order to enable the bit error to decay faster as block length increases, the encoder has to code information bits \emph{jointly}. A celebrated feedback coding scheme, Shalkwijk--Kailath scheme, simplified/illustrated in Figure~\ref{KSenc}, belongs to this category.

\noindent {\bf S-K scheme.} Here all information bits are used only to generate the first codeword. The rest of the codewords depend only on the feedback (noise added to the previous transmission). Although S-K scheme does encode all information bits jointly, transmitting all information bits in the first phase requires a high numerical precision as block length increases. For example, for $50$ information bits, the transmitter transmits $\sum_{k=1}^K b_k 2^k$ (with a power normalization and subtracting a constant to set mean to be 0).

\begin{figure}[!ht] 
\centering
\includegraphics[width=0.5\textwidth]{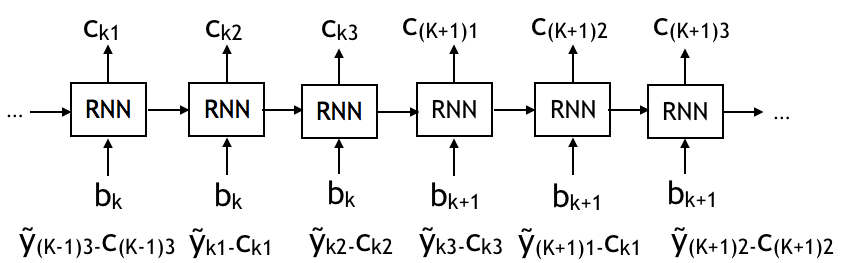}
\caption{Bit-coupled RNN encoder}\label{simple1}
\end{figure}

Our approach is different from S-K scheme in that we aim to use the memory of RNN to design an encoder that encodes the information bits jointly. 
Since RNN has a memory in it, naturally it allows encoding bits jointly. The challenge is whether we can find/train a neural network encoder which makes use of the RNN memory. 
For example, Figure~\ref{simple1} illustrates a somewhat natural architecture we attempted. However, after training, the BER performance is only as good as the BER of bit-by-bit encoding, which means that the memory in the RNN is not being successfully used to jointly encode the information bits. 

%

%
%
%
%
%
%

\section{Robustness under bursty Gaussian channels}\label{app:robustness}


Bursty Gaussian channel is a channel where there is a background Gaussian noise $n_{1i}$, and occasionally, with a small probability $\alpha$, a Gaussian noise with high power (bursty noise, $n_{2i}$) is added on top of the background noise.
Mathematically, we consider the following bursty Gaussian channel:
$y_i = x_i + n_i$, where
\begin{align*}
n_i &= n_{1i} + e_i n_{2i}, \\
n_{1i} &\sim \mathcal{N}(0,\sigma_o^2), \ \ n_{2i} \sim \mathcal{N}(0,\sigma_1^2), \ \ e_i \sim Bern(\alpha).
\end{align*}

\begin{figure}[!ht]
\centering
\includegraphics[width=0.5\textwidth]{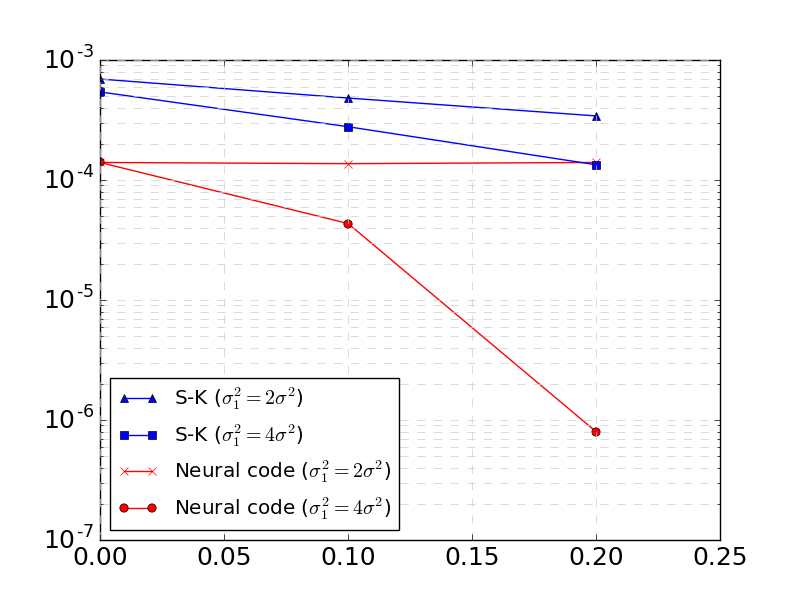}
	\put(-260,120){BER}
	\put(-186,-3){Probability of burst noise ($\alpha$)}
\caption{Neural feedback code is robust to bursty noise.}\label{fig:A1_BER}

\end{figure}

We test the robustness of our feedback code under bursty Gaussian channel. 
Figure~\ref{fig:A1_BER} shows the BER as a function of $\alpha$ (probability of having a burst noise), for $-1dB$, and two different power of burst noise.
We choose $\sigma_0^2$ so that $\alpha \sigma_1^2 + \sigma_0^2 = \sigma^2$ (i.e., we keep the total power of the noise). As we can see from the figure, as $\alpha$ increases, the BER decreases, showing that the bit error rate is smaller for more bursty noise channels.

\end{document}